\documentclass[10pt,journal,epsfig]{IEEEtran}

\usepackage[dvips]{graphicx}
\usepackage{graphicx}
\usepackage{amssymb}
\usepackage{cite}
\usepackage{subfigure}
\usepackage{amsmath}
\usepackage{multirow}
\usepackage{booktabs}

\begin{document}

\title{Sparse Bayesian Dictionary Learning with a Gaussian Hierarchical Model}

\author{Linxiao Yang, Jun Fang, Hong Cheng, and Hongbin Li,~\IEEEmembership{Senior
Member,~IEEE}
\thanks{Linxiao Yang, and Jun Fang are with the National Key Laboratory on Communications,
University of Electronic Science and Technology of China, Chengdu
611731, China, Email: JunFang@uestc.edu.cn}
\thanks{Hong Cheng is with the School of Automation, University of Electronic Science and Technology of China,
Chengdu 611731, China, Email: hcheng@uestc.edu.cn}
\thanks{Hongbin Li is
with the Department of Electrical and Computer Engineering,
Stevens Institute of Technology, Hoboken, NJ 07030, USA, E-mail:
Hongbin.Li@stevens.edu}
\thanks{This work was supported in part by the National Science
Foundation of China under Grant 61172114, and the National Science
Foundation under Grant ECCS-1408182. }}

%\thanks{Huiping Duan is with the School of Electronic Engineering,
%University of Electronic Science and Technology of China, Chengdu
%611731, China, Email: huipingduan@uestc.edu.cn}

\maketitle

\begin{abstract}
We consider a dictionary learning problem whose objective is to
design a dictionary such that the signals admits a sparse or an
approximate sparse representation over the learned dictionary.
Such a problem finds a variety of applications such as image
denoising, feature extraction, etc. In this paper, we propose a
new hierarchical Bayesian model for dictionary learning, in which
a Gaussian-inverse Gamma hierarchical prior is used to promote the
sparsity of the representation. Suitable priors are also placed on
the dictionary and the noise variance such that they can be
reasonably inferred from the data. Based on the hierarchical
model, a variational Bayesian method and a Gibbs sampling method
are developed for Bayesian inference. The proposed methods have
the advantage that they do not require the knowledge of the noise
variance \emph{a priori}. Numerical results show that the proposed
methods are able to learn the dictionary with an accuracy better
than existing methods, particularly for the case where there is a
limited number of training signals.
\end{abstract}


%``off-the-shelf''
%instead of a prespecified
%and can better fit the data

\section{Introduction}
Sparse representation has been of significant interest over past
few years and has found a variety of applications in practice as
many natural signals admit a sparse or an approximate sparse
representation in a certain basis
\cite{CandesTao05,CarvajalinoSapiro09,WrightYang09}. In many
applications such as image denoising and interpolation, signals
are assumed to admit a sparse representation over a pre-specified
non-adaptive dictionary, e.g. discrete consine/wavelet transform
(DCT/DWT) bases. Nevertheless, recent research
\cite{AharonElad06,EladAharon06} has shown that the recovery,
denoising and classification performance can be considerably
improved by utilizing an adaptive dictionary that is learned from
the training signals \cite{EladAharon06,MairalBach12}. This has
inspired studies on dictionary learning whose objective is to
design overcompelete dictionaries that can better represent the
signals. A number of algorithms, such as K-singular value
decomposition (K-SVD) \cite{AharonElad06}, method of optimal
directions (MOD) \cite{EnganAase99}, dictionary learning with the
majorization method \cite{YaghoobiBlumensath09}, and simultaneous
codeword optimization (SimCO) \cite{DaiXu12}, were developed for
learning overcomplete dictionaries for sparse representation. Most
algorithms formulate the dictionary learning as an optimization
problem and solve it via a two-stage iterative process, namely, a
sparse coding stage and a dictionary update stage. The main
difference between these algorithms lies in the dictionary update
stage. Specifically, the MOD method \cite{EnganAase99} updates the
dictionary via solving a least square problem which admits a
closed-form for the dictionary update. The K-SVD algorithm
\cite{AharonElad06}, instead, updates atoms of the dictionary in a
sequential manner and while updating each atom, the atom is
updated along with the nonzero entries in the corresponding row
vector of the sparse matrix. The idea of this sequential atom
update was later extended to sequentially updating multiple atoms
each time \cite{DaiXu12}, and recently was generalized to parallel
atom-updating in order to further accelerate the convergence of
the iterative process \cite{SadeghiZadeh14}. These methods
\cite{AharonElad06,EnganAase99,YaghoobiBlumensath09,DaiXu12,SadeghiZadeh14},
although delivering state-of-the-art performance, require the
knowledge of the sparsity level or the noise/residual variance to
define the stopping criterion for estimating the sparse codes
(e.g. \cite{AharonElad06}), or select appropriate values for the
regularization parameters controlling the tradeoff between the
sparsity level and the data fitting error (e.g.
\cite{YaghoobiBlumensath09,SadeghiZadeh14}). In practice, however,
the prior information about the noise variance is usually
unavailable and an inaccurate estimation may result in substantial
performance degradation. To mitigate this limitation, a
nonparametric Bayesian dictionary learning method called as
beta-Bernoulli process factor analysis (BPFA) was recently
developed in \cite{ZhouChen12}. The proposed method is able to
automatically infer the required number of factors (dictionary
elements) and the noise variance from the image under test, which
is deemed as an important advantage over other dictionary learning
methods. For \cite{ZhouChen12}, the posterior distributions cannot
be derived analytically, and a Gibbs sampler was used for Bayesian
inference. We also note that a class of online dictionary learning
algorithms were developed in
\cite{MairalBach10,SkrettingEngan10,LabuschBarth11}. Different
from the above batch-based algorithms
\cite{AharonElad06,EnganAase99,DaiXu12,SadeghiZadeh14} which use
the whole set of training data for dictionary learning, online
algorithms continuously update the dictionary using only one or a
small batch of training data, which enables them to handle very
large data sets.

%in which a beta-Bernoulli prior was employed to induce sparse
%representations

%employed to induce sparse representations

%This joint atom and sparse signal update results in a fast
%convergence rate and meanwhile helps achieve quite decent
%performance.

%along with their corresponding rows of the sparse matrix

%lead to sparse representations of signals.

%The K-SVD is a widely used algorithm with great success for
%dictionary learning.

%that has been widely used in compressed sensing

%for learning overcomplete dictionaries for sparse representation

%which can infer the noise variance and the sparsity level
%automatically from the data

In this paper, we propose a new hierarchical Bayesian model for
dictionary learning, in which a Gaussian-inverse Gamma
hierarchical prior is used to promote the sparsity of the
representation. Suitable priors are also placed on the dictionary
and the noise variance such that they can be reasonably inferred
from the data. Based on the hierarchical model, a variational
Bayesian method and a Gibbs sampling method are developed for
Bayesian inference. For both inference methods, there are two
different ways to update the dictionary: we can update the whole
set of atoms at once, or update the atoms in a sequential manner.
When updating the dictionary as a whole, the proposed variational
Bayesian method has a dictionary update formula similar to the MOD
method. Nevertheless, unlike the MOD method which alternates
between two separate stages (i.e. dictionary update and sparse
coding), for our algorithm, the dictionary and the signal are
refined in an interweaved and gradual manner, which enables the
algorithm to come to a reasonably nearby point as the optimization
progresses, and helps avoid undesirable local minima. For the
Gibbs sampler, a sequential update seems able to expedite the
convergence rate and helps achieve better performance. Simulation
results show that the proposed Gibbs sampling algorithm presents
uniform superiority over other state-of-the-art dictionary
learning methods in a number of experiments.

The rest of the paper is organized as follows. In Section
\ref{sec:model}, we introduce a hierarchical prior model for
learning dictionaries. Based on this hierarchical model, a
variational Bayesian method and a Gibbs sampler are developed in
Section \ref{sec:VB} and Section \ref{sec:Gibbs} for Bayesian
inference. Simulation results are provided in Section
\ref{sec:simulation}, followed by concluding remarks in Section
\ref{sec:conclusion}.

%Note that our proposed hierarchical model is flexible to
%accommodate additional prior information,

%has demonstrated superior results in denoising applications

%The proposed algorithms no longer need the knowledge of the noise
%variance

%in order to prevent the dictionary from becoming infinitely large

%The Gaussian-inverse Gamma prior is a popular sparse-promoting
%prior that has been widely used in compressed sensing
%\cite{WipfRao07,JiXue08,FangShen15}. As discussed in
%\cite{Tipping01}, this two-layer Gaussian-inverse Gamma
%hierarchical prior model tends to switch off most of the
%coefficients that are deemed to be irrelevant, and only keep very
%few relevant coefficients to explain the data. This mechanism,
%also called as ``automatic relevance determination'',

\section{Hierarchical Model} \label{sec:model}
Suppose we have $L$ training signals
$\{\boldsymbol{y}_l\}_{l=1}^L$, where
$\boldsymbol{y}_l\in\mathbb{R}^M$. Dictionary learning aims at
finding a common sparsifying dictionary
$\boldsymbol{D}\in\mathbb{R}^{M\times N}$ such that these $L$
training signals admit a sparse representation over the
overcomplete dictionary $\boldsymbol{D}$, i.e.
\begin{align}
\boldsymbol{y}_l=\boldsymbol{D}\boldsymbol{x}_l+\boldsymbol{w}_l
\qquad \forall l \label{model-1}
\end{align}
where $\boldsymbol{x}_l$ and $\boldsymbol{w}_l$ denote the sparse
vector and the residual/noise vector, respectively. Define
$\boldsymbol{Y}\triangleq
[\boldsymbol{y}_1\phantom{0}\ldots\phantom{0}\boldsymbol{y}_L]$,
$\boldsymbol{X}\triangleq
[\boldsymbol{x}_1\phantom{0}\ldots\phantom{0}\boldsymbol{x}_L]$,
and $\boldsymbol{W}\triangleq
[\boldsymbol{w}_1\phantom{0}\ldots\phantom{0}\boldsymbol{w}_L]$,
the model (\ref{model-1}) can be re-expressed as
\begin{align}
\boldsymbol{Y}=\boldsymbol{D}\boldsymbol{X}+\boldsymbol{W}
\end{align}
Also, we write $\boldsymbol{D}\triangleq
[\boldsymbol{d}_1\phantom{0}\ldots\phantom{0}\boldsymbol{d}_N]$,
where each column of the dictionary, $\boldsymbol{d}_n$, is called
an atom.

%We wish to infer a common dictionary $\boldsymbol{D}$ and the
%sparse matrix $\boldsymbol{X}$ simultaneously from the observed
%data $\boldsymbol{Y}$. To this objective

%and the parameters $a$ and $b$ used to characterize the Gamma
%distribution are chosen to be very small values, e.g. $10^{-4}$,
%in order to provide non-informative/uniform (over a logarithmic
%scale) hyperpriors over $\{\alpha_{nl}\}$.

In the following, we develop a Bayesian framework for learning the
overcomplete dictionary and sparse vectors. To promote sparse
representations, we assign a two-layer hierarchical
Gaussian-inverse Gamma prior to $\boldsymbol{X}$. The
Gaussian-inverse Gamma prior is one of the most popular
sparse-promoting priors which has been widely used in compressed
sensing \cite{WipfRao07,JiXue08,FangShen15}. In the first layer,
$\boldsymbol{X}$ is assigned a Gaussian prior distribution
\begin{align}
p(\boldsymbol{X}|\boldsymbol{\alpha})=&\prod_{n=1}^N\prod_{l=1}^L
p(x_{nl}) \nonumber\\
=&\prod_{n=1}^N\prod_{l=1}^L
\mathcal{N}(x_{nl}|0,\alpha_{nl}^{-1})
\end{align}
where $x_{nl}$ denotes the $(n,l)$th entry of $\boldsymbol{X}$,
and $\boldsymbol{\alpha}\triangleq\{\alpha_{nl}\}$ are
non-negative sparsity-controlling hyperparameters. The second
layer specifies Gamma distributions as hyperpriors over the
hyperparameters $\{\alpha_{nl}\}$, i.e.
\begin{align}
p(\boldsymbol{\alpha})=&\prod_{n=1}^N\prod_{l=1}^L
\text{Gamma}(\alpha_{nl}|a,b) \nonumber\\
=&
\prod_{n=1}^N\prod_{l=1}^L\Gamma(a)^{-1}b^{a}\alpha_{nl}^{a-1}e^{-b\alpha_{nl}}
\end{align}
where $\Gamma(a)=\int_{0}^{\infty}t^{a-1}e^{-t}dt$ is the Gamma
function, and the parameters $a$ and $b$ used to characterize the
Gamma distribution are usually chosen to be small values, e.g.
$10^{-6}$. As discussed in \cite{Tipping01}, this hyperprior
allows the posterior mean of $\alpha_{nl}$ to become arbitrarily
large. As a consequence, the associated coefficient $x_{nl}$ will
be driven to zero, thus yielding a sparse solution. In this paper,
we choose a value of $a=0.5$ in order to achieve a more
sparsity-encouraging effect. Clearly, the Gamma prior with a
larger $a$ encourages large values of the hyperparameters, and
therefore promotes the sparseness of the solution since the larger
the hyperparameter, the smaller the variance of the corresponding
coefficient.

\begin{figure}[!t]
\centering
\includegraphics[width=5cm]{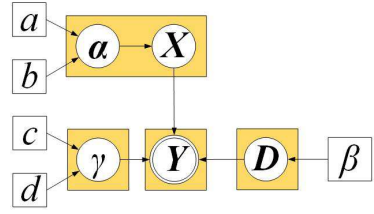}
\caption{Hierarchical model for dictionary learning.}
\label{fig:model}
\end{figure}

In addition, in order to prevent the dictionary from becoming
infinitely large, we assume the atoms of the dictionary
$\{\boldsymbol{d}_n\}$ are mutually independent and each atom is
placed a Gaussian prior, i.e.
\begin{align}
p(\boldsymbol{D})=\prod_{n=1}^N p(\boldsymbol{d}_n)=\prod_{n=1}^N
\mathcal{N}(\boldsymbol{d}_n|\boldsymbol{0},\beta\boldsymbol{I})
\end{align}
where $\beta$ is a parameter whose choice will be discussed later.
The noise $\{\boldsymbol{w}_l\}$ are assumed independent
multivariate Gaussian noise with zero mean and covariance matrix
$(1/\gamma)\boldsymbol{I}$, where the noise variance $1/\gamma$ is
assumed unknown \emph{a priori}. To estimate the noise variance,
we place a Gamma hyperprior over $\gamma$, i.e.
\begin{align}
p(\gamma)=\text{Gamma}(\gamma|c,d)=\Gamma(c)^{-1}d^{c}\gamma^{c-1}e^{-d\gamma}
\label{gamma-prior}
\end{align}
where we set $c=0.5$ and $d=10^{-6}$. The proposed hierarchical
model (see Fig. \ref{fig:model}) provides a general framework for
learning the overcomplete dictionary, the sparse codes, as well as
the noise variance. In the following, we will develop a
variational Beyesian method and a Gibbs sampling method for
Bayesian inference.

%The parameter $b$ is usually chosen to be a very small value, say,
%$b=10^{-6}$. The choice of $a$ is less critical and we set $a=0.5$
%in our paper.
%in order to prevent the dictionary from becoming infinitely large

\section{Variational Inference} \label{sec:VB}
\subsection{Review of The Variational Bayesian Methodology}
Before proceeding, we firstly provide a brief review of the
variational Bayesian methodology. In a probabilistic model, let
$\boldsymbol{y}$ and $\boldsymbol{\theta}$ denote the observed
data and the hidden variables, respectively. It is straightforward
to show that the marginal probability of the observed data can be
decomposed into two terms
\begin{align}
\ln p(\boldsymbol{y})=L(q)+\text{KL}(q|| p)
\label{variational-decomposition}
\end{align}
where
\begin{align}
L(q)=\int q(\boldsymbol{\theta})\ln
\frac{p(\boldsymbol{y},\boldsymbol{\theta})}{q(\boldsymbol{\theta})}d\boldsymbol{\theta}
\end{align}
and
\begin{align}
\text{KL}(q|| p)=-\int q(\boldsymbol{\theta})\ln
\frac{p(\boldsymbol{\theta}|\boldsymbol{y})}{q(\boldsymbol{\theta})}d\boldsymbol{\theta}
\end{align}
where $q(\boldsymbol{\theta})$ is any probability density
function, $\text{KL}(q|| p)$ is the Kullback-Leibler divergence
between $p(\boldsymbol{\theta}|\boldsymbol{y})$ and
$q(\boldsymbol{\theta})$. Since $\text{KL}(q|| p)\geq 0$, it
follows that $L(q)$ is a rigorous lower bound on $\ln
p(\boldsymbol{y})$. Moreover, notice that the left hand side of
(\ref{variational-decomposition}) is independent of
$q(\boldsymbol{\theta})$. Therefore maximizing $L(q)$ is
equivalent to minimizing $\text{KL}(q|| p)$, and thus the
posterior distribution $p(\boldsymbol{\theta}|\boldsymbol{y})$ can
be approximated by $q(\boldsymbol{\theta})$ through maximizing
$L(q)$.

%by assuming an appropriate $q(\boldsymbol{\theta})$

The significance of the above transformation is that it
circumvents the difficulty of computing the posterior probability
$p(\boldsymbol{\theta}|\boldsymbol{y})$ (which is usually
computationally intractable). For a suitable choice for the
distribution $q(\boldsymbol{\theta})$, the quantity $L(q)$ may be
more amiable to compute. Specifically, we could assume some
specific parameterized functional form for
$q(\boldsymbol{\theta})$ and then maximize $L(q)$ with respect to
the parameters of the distribution. A particular form of
$q(\boldsymbol{\theta})$ that has been widely used with great
success is the factorized form over the component variables
$\{\theta_i\}$ in $\boldsymbol{\theta}$ \cite{TzikasLikas08}, i.e.
$q(\boldsymbol{\theta})=\prod_i q_i(\theta_i)$. We therefore can
compute the posterior distribution approximation by finding
$q(\boldsymbol{\theta})$ of the factorized form that maximizes the
lower bound $L(q)$. The maximization can be conducted in an
alternating fashion for each latent variable, which leads to
\cite{TzikasLikas08}
\begin{align}
q_i(\theta_i)=\frac{\exp(\langle\ln
p(\boldsymbol{y},\boldsymbol{\theta})\rangle_{k\neq
i})}{\int\exp(\langle\ln
p(\boldsymbol{t},\boldsymbol{\theta})\rangle_{k\neq i})d\theta_i}
\label{general-update}
\end{align}
where $\langle\cdot\rangle_{k\neq i}$ denotes an expectation with
respect to the distributions $q_i(\theta_i)$ for all $k\neq i$.

\subsection{Proposed Variational Bayesian Method}
We now proceed to perform variational Bayesian inference for the
proposed hierarchical model. Let
$\boldsymbol{\theta}\triangleq\{\boldsymbol{X},\boldsymbol{\alpha},
\boldsymbol{D},\gamma\}$ denote all hidden variables. We assume
posterior independence among the variables $\boldsymbol{X}$,
$\boldsymbol{\alpha}$, $\boldsymbol{D}$ and $\gamma$, i.e.
\begin{align}
p(\boldsymbol{\theta}|\boldsymbol{y})\approx &
q(\boldsymbol{x},\boldsymbol{\alpha},\boldsymbol{D},\gamma)\nonumber\\
=&
q_x(\boldsymbol{x})q_{\alpha}(\boldsymbol{\alpha})q_{d}(\boldsymbol{D})q_{\gamma}(\gamma)
\end{align}
With this mean field approximation, the posterior distribution of
each hidden variable can be computed by maximizing $L(q)$ while
keeping other variables fixed using their most recent
distributions, which gives
\begin{align}
\ln q_x(\boldsymbol{X})=&\langle\ln
p(\boldsymbol{Y},\boldsymbol{X},\boldsymbol{D},\boldsymbol{\alpha},\gamma)\rangle_{q_{d}(\boldsymbol{D})
q_{\alpha}(\boldsymbol{\alpha})
q_{\gamma}(\gamma)} + \text{constant} \nonumber\\
\ln q_d(\boldsymbol{D})=&\langle\ln
p(\boldsymbol{Y},\boldsymbol{X},\boldsymbol{D},\boldsymbol{\alpha},\gamma)\rangle_{q_{x}(\boldsymbol{X})
q_{\alpha}(\boldsymbol{\alpha})
q_{\gamma}(\gamma)} + \text{constant} \nonumber\\
\ln q_{\alpha}(\boldsymbol{\alpha})=&\langle\ln
p(\boldsymbol{Y},\boldsymbol{X},\boldsymbol{D},\boldsymbol{\alpha},\gamma)\rangle_{q_x(\boldsymbol{X})
q_{d}(\boldsymbol{D})q_{\gamma}(\gamma)} + \text{constant} \nonumber\\
\ln q_{\gamma}(\gamma)=&\langle\ln
p(\boldsymbol{Y},\boldsymbol{X},\boldsymbol{D},\boldsymbol{\alpha},\gamma)\rangle_{q_x(\boldsymbol{X})
q_{d}(\boldsymbol{D})q_{\alpha}(\boldsymbol{\alpha})} +
\text{constant} \nonumber
\end{align}
where $\langle\rangle_{q_1(\cdot)\ldots q_K(\cdot)}$ denotes the
expectation with respect to (w.r.t.) the distributions
$\{q_k(\cdot)\}_{k=1}^K$. In summary, the posterior distribution
approximations are computed in an alternating fashion for each
hidden variable, with other variables fixed. Details of this
Bayesian inference scheme are provided below.

%For simplicity, $\boldsymbol{X}$ can be updated in a columnwise
%manner, that is,

\textbf{\emph{1). Update of $q_x(\boldsymbol{X})$}}: The
calculation of $q_x(\boldsymbol{X})$ can be decomposed into a set
of independent tasks, with each task computing the posterior
distribution approximation for each column of $\boldsymbol{X}$,
i.e. $q_{x}(\boldsymbol{x}_l)$. We have
\begin{align}
\ln q_{x}(\boldsymbol{x}_l)\propto & \langle\ln
[p(\boldsymbol{y}_l|\boldsymbol{D},\boldsymbol{x}_l,\gamma)p(\boldsymbol{x}_l|\boldsymbol{\alpha}_l)]
\rangle_{q_{d}(\boldsymbol{D}) q_{\alpha}(\boldsymbol{\alpha})
q_{\gamma}(\gamma)} \label{eqn-2}
\end{align}
where $\boldsymbol{\alpha}_l\triangleq\{\alpha_{nl}\}_{n=1}^N$ is
the sparsity-controlling hyperparameters associated with
$\boldsymbol{x}_l$,
$p(\boldsymbol{y}_l|\boldsymbol{D},\boldsymbol{x}_l,\gamma)$ and
$p(\boldsymbol{x}_l|\boldsymbol{\alpha}_l)$ are respectively given
by
\begin{align}
p(\boldsymbol{y}_l|\boldsymbol{D},\boldsymbol{x}_l,\gamma)=&\left(\frac{\gamma}{2\pi}\right)^{\frac{M}{2}}
\exp\left(-\frac{\gamma\|\boldsymbol{y}_l-\boldsymbol{D}\boldsymbol{x}_l\|_2^2}{2}\right)
\nonumber\\
p(\boldsymbol{x}_l|\boldsymbol{\alpha}_l)=&\prod_{n=1}^N
\mathcal{N}(x_{nl}|0,\alpha_{nl}^{-1}) \label{eqn-1}
\end{align}
Substituting (\ref{eqn-1}) into (\ref{eqn-2}) and after some
simplifications, it can be readily verified that
$q_x(\boldsymbol{x}_l)$ follows a Gaussian distribution
\begin{align}
q_x(\boldsymbol{x}_l)=\mathcal{N}(\boldsymbol{x}_l|\boldsymbol{\mu}_l^x,\boldsymbol{\Sigma}_l^x)
\label{x-update}
\end{align}
with its mean $\boldsymbol{\mu}_l^x$ and covariance matrix
$\boldsymbol{\Sigma}_l^x$ given respectively as
\begin{align}
\boldsymbol{\mu}_l^x=&\langle\gamma\rangle\boldsymbol{\Sigma}_l^x\langle\boldsymbol{D}\rangle^T\boldsymbol{y}_l
\nonumber\\
\boldsymbol{\Sigma}_l^x=&\left(\langle\gamma\rangle\langle\boldsymbol{D}^T
\boldsymbol{D}\rangle+\langle\boldsymbol{\Lambda}_l\rangle\right)^{-1}
\end{align}
where $\langle\gamma\rangle$ denotes the expectation w.r.t.
$q_{\gamma}(\gamma)$, $\langle\boldsymbol{D}\rangle$ and
$\langle\boldsymbol{D}^T\boldsymbol{D}\rangle$ denote the
expectation w.r.t. $q_{d}(\boldsymbol{D})$, and
$\langle\boldsymbol{\Lambda}_l\rangle\triangleq\text{diag}(\langle
\alpha_{1l}\rangle, \ldots, \langle \alpha_{Nl}\rangle)$, in which
$\langle \alpha_{nl}\rangle$ represents the expectation w.r.t.
$q_{\alpha}(\boldsymbol{\alpha})$.

%for notational simplicity, we omit the subscript for the
%$\langle\cdot\rangle$ operation. To

\textbf{\emph{2). Update of $q_d(\boldsymbol{D})$}}: The
approximate posterior $q_{d}(\boldsymbol{D})$ can be obtained as
\begin{align}
\ln q_d(\boldsymbol{D})\propto&\langle\ln
[p(\boldsymbol{Y}|\boldsymbol{X},\boldsymbol{D},\gamma)p(\boldsymbol{D})]\rangle_{q_x(\boldsymbol{X})
q_{\gamma}(\gamma)} \nonumber\\
\propto& \langle -\gamma
\|\boldsymbol{Y}-\boldsymbol{D}\boldsymbol{X}\|_{F}^2
-\beta^{-1}\sum_{n=1}^N \boldsymbol{d}_n^T\boldsymbol{d}_n\rangle
\nonumber\\
\propto& \langle
-\gamma\text{tr}\{(\boldsymbol{Y}-\boldsymbol{D}\boldsymbol{X})
(\boldsymbol{Y}-\boldsymbol{D}\boldsymbol{X})^T\}-\beta^{-1}\text{tr}\{\boldsymbol{D}\boldsymbol{D}^T
\}\rangle \nonumber\\
\propto& \langle
\text{tr}\{\boldsymbol{D}(\gamma\boldsymbol{X}\boldsymbol{X}^T+\beta^{-1}\boldsymbol{I})
\boldsymbol{D}^T-2\gamma\boldsymbol{Y}\boldsymbol{X}^T\boldsymbol{D}^T\}\rangle
\nonumber\\
=&
\text{tr}\{\boldsymbol{D}(\langle\gamma\rangle\langle\boldsymbol{X}\boldsymbol{X}^T\rangle+\beta^{-1}\boldsymbol{I})
\boldsymbol{D}^T-2\langle\gamma\rangle\boldsymbol{Y}\langle\boldsymbol{X}\rangle^T\boldsymbol{D}^T\}
\end{align}
where for simplicity, we have dropped the subscript of the
$\langle\cdot\rangle$ operator. Define
\begin{align}
\boldsymbol{A}\triangleq &
(\langle\gamma\rangle\langle\boldsymbol{X}\boldsymbol{X}^T\rangle+\beta^{-1}\boldsymbol{I})^{-1}
\nonumber\\
\boldsymbol{B}\triangleq &
\langle\gamma\rangle\boldsymbol{Y}\langle\boldsymbol{X}\rangle^T
\nonumber
\end{align}
The posterior $q_{d}(\boldsymbol{D})$ can be further expressed as
\begin{align}
\ln
q_d(\boldsymbol{D})\propto&\text{tr}\{\boldsymbol{D}\boldsymbol{A}^{-1}\boldsymbol{D}^T-2\boldsymbol{B}\boldsymbol{D}^T\}
\nonumber\\
=&\sum_{m=1}^M(\boldsymbol{d}_{m\cdot}\boldsymbol{A}^{-1}\boldsymbol{d}_{m\cdot}^T-2\boldsymbol{b}_{m\cdot}
\boldsymbol{d}_{m\cdot}^T) \label{eqn-3}
\end{align}
where $\boldsymbol{b}_{m\cdot}$ and $\boldsymbol{d}_{m\cdot}$
represents the $m$th row of $\boldsymbol{B}$ and $\boldsymbol{D}$,
respectively. It can be easily seen from (\ref{eqn-3}) that the
posterior distribution $q_d(\boldsymbol{D})$ has independent rows
and each row follows a Gaussian distribution with its mean and
covariance matrix given by $\boldsymbol{b}_{m\cdot}\boldsymbol{A}$
and $\boldsymbol{A}$, respectively, i.e.
\begin{align}
q_d(\boldsymbol{D})=\prod_{m=1}^M
p(\boldsymbol{d}_{m\cdot})=\prod_{m=1}^M
\mathcal{N}(\boldsymbol{b}_{m\cdot}\boldsymbol{A},\boldsymbol{A})
\label{D-update}
\end{align}

%with its mean and covariance matrix given by
%\begin{align}
%\boldsymbol{d}_{i\cdot}\sim
%\mathcal{N}(\boldsymbol{b}_{i\cdot}\boldsymbol{A},\boldsymbol{A})
%\label{D-update}
%\end{align}

%Let $\boldsymbol{\vec{d}}\triangleq\text{vec}(\boldsymbol{D}^T)$
%denote the vectorization of the matrix $\boldsymbol{D}^T$ formed
%by stacking its column vectors. Then we can easily show that
%\begin{align}
%\boldsymbol{\vec{d}}\sim
%\mathcal{N}(\boldsymbol{\mu}_d,\boldsymbol{\Sigma}_d)
%\end{align}
%where $\boldsymbol{\mu}_d$ and $\boldsymbol{\Sigma}_d$ are
%respectively given as
%\begin{align}
%\boldsymbol{\mu}_d=&(\text{vec}(\boldsymbol{B}^T))^T\boldsymbol{\Sigma}_d
%\nonumber\\
%\boldsymbol{\Sigma}_d=&\boldsymbol{I}\otimes\boldsymbol{A}
%\nonumber
%\end{align}
%in which $\otimes$ denotes the kronecker product.

%where $\langle x_{ij}^2\rangle$ denotes the expectation w.r.t.
%$q_x(\boldsymbol{X})$.

\textbf{\emph{3). Update of $q_{\alpha}(\boldsymbol{\alpha})$}}:
The variational optimization of $q_{\alpha}(\boldsymbol{\alpha})$
yields
\begin{align}
\ln q_{\alpha}(\boldsymbol{\alpha})\propto&\langle\ln
p(\boldsymbol{X}|\boldsymbol{\alpha})p(\boldsymbol{\alpha})\rangle_{q_x(\boldsymbol{X})}
\nonumber\\
=&\sum_{n=1}^N\sum_{l=1}^L \langle\ln
p(x_{nl}|\alpha_{nl})p(\alpha_{nl}; a,b)\rangle \nonumber\\
\propto & \sum_{n=1}^N\sum_{l=1}^L\left\{
\left(a-\frac{1}{2}\right)\ln\alpha_{nl}-\left(b+\frac{\langle
x_{nl}^2\rangle}{2}\right)\alpha_{nl} \right\}
\end{align}
Thus $\boldsymbol{\alpha}$ has a form of a product of Gamma
distributions
\begin{align}
q_{\alpha}(\boldsymbol{\alpha})=\prod_{n=1}^N\prod_{l=1}^L
\text{Gamma}(\alpha_{nl}; \tilde{a}, \tilde{b}_{nl})
\label{alpha-update}
\end{align}
in which the parameters $\tilde{a}$ and $\tilde{b}_{nl}$ are
respectively given as
\begin{align}
\tilde{a}=a+\frac{1}{2} \qquad \tilde{b}_{nl}=b+\frac{1}{2}\langle
x_{nl}^2\rangle
\end{align}

\textbf{\emph{4). Update of $q_{\gamma}(\gamma)$}}: The
variational optimization of $q_{\gamma}(\gamma)$ yields
\begin{align}
\ln q_{\gamma}(\gamma)\propto&\langle\ln
p(\boldsymbol{Y}|\boldsymbol{D},\boldsymbol{X},\gamma)p(\gamma)\rangle_{q_d(\boldsymbol{D})q_x(\boldsymbol{X})}
\nonumber\\
\propto&\langle\ln\prod_{l=1}^L
p(\boldsymbol{y}_l|\boldsymbol{D},\boldsymbol{x}_l,\gamma)p(\gamma)\rangle
\nonumber\\
\propto&
\langle\frac{ML}{2}\ln\gamma-\frac{\gamma}{2}\sum_{l=1}^L(\boldsymbol{y}_l-\boldsymbol{D}\boldsymbol{x}_l)^T
(\boldsymbol{y}_l-\boldsymbol{D}\boldsymbol{x}_l) \nonumber\\
& +(c-1)\ln\gamma-d\gamma\rangle \nonumber\\
=&\bigg(\frac{ML}{2}+c-1\bigg)\ln\gamma-
\bigg(\frac{1}{2}\langle\|\boldsymbol{Y}-\boldsymbol{D}\boldsymbol{X}\|_F^2\rangle+d\bigg)\gamma
\end{align}
Therefore $q_{\gamma}(\gamma)$ follows a Gamma distribution
\begin{align}
q_{\gamma}(\gamma)=\text{Gamma}(\gamma|\tilde{c},\tilde{d})
\label{gamma-update}
\end{align}
with the parameters $\tilde{c}$ and $\tilde{d}$ given respectively
by
\begin{align}
\tilde{c}=&\frac{ML}{2}+c \nonumber\\
\tilde{d}=&d+\frac{1}{2}\langle\|\boldsymbol{Y}-\boldsymbol{D}\boldsymbol{X}\|_F^2
\rangle
\end{align}
where
\begin{align}
\langle\|\boldsymbol{Y}-\boldsymbol{D}\boldsymbol{X}\|_F^2 \rangle
=&\langle\text{tr}\{(\boldsymbol{Y}-\boldsymbol{D}\boldsymbol{X})^T
(\boldsymbol{Y}-\boldsymbol{D}\boldsymbol{X})\}\rangle \nonumber\\
=&\|\boldsymbol{Y}-\langle\boldsymbol{D}\rangle\langle\boldsymbol{X}\rangle\|_F^2+
\text{tr}\{\langle\boldsymbol{D}^T\boldsymbol{D}\rangle\langle\boldsymbol{X}\boldsymbol{X}^T\rangle\}
\nonumber\\
&-\text{tr}\{\langle\boldsymbol{D}^T\rangle\langle\boldsymbol{D}\rangle\langle\boldsymbol{X}\rangle
\langle\boldsymbol{X}^T\rangle\}
\end{align}

In summary, the variational Bayesian inference involves updates of
the approximate posterior distributions for hidden variables
$\boldsymbol{X}$, $\boldsymbol{D}$, $\boldsymbol{\alpha}$, and
$\gamma$. Some of the expectations and moments used during the
update are summarized as
\begin{align}
\langle
x_{nl}^2\rangle\stackrel{(a)}{=}&(\boldsymbol{\mu}_{l}^{x}[n])^2+\boldsymbol{\Sigma}_l^{x}[n,n]
\nonumber\\
\langle\boldsymbol{X}\boldsymbol{X}^T\rangle=&\langle\boldsymbol{X}\rangle\langle\boldsymbol{X}\rangle^T
+\sum_{l=1}^L\boldsymbol{\Sigma}_l^{x} \nonumber\\
\langle\boldsymbol{D}\rangle\stackrel{(b)}{=}&\boldsymbol{B}\boldsymbol{A}
\nonumber\\
\langle\boldsymbol{D}^T\boldsymbol{D}\rangle=&\langle\boldsymbol{D}\rangle^T
\langle\boldsymbol{D}\rangle+M\langle\boldsymbol{A}\rangle
\nonumber\\
\langle\alpha_{nl}\rangle=&\tilde{a}/\tilde{b}_{nl}
\nonumber\\
\langle\gamma\rangle=&\tilde{c}/\tilde{d} \nonumber
\end{align}
where in $(a)$, $\boldsymbol{\mu}_{l}^{x}[n]$ denotes the $n$th
entry of $\boldsymbol{\mu}_{l}^{x}$,
$\boldsymbol{\Sigma}_l^{x}[n,n]$ represents the $n$th diagonal
element of $\boldsymbol{\Sigma}_l^{x}$, and $(b)$ follows from
(\ref{D-update}). For clarity, we summarize our algorithm as
follows.

\begin{center}
\textbf{Sparse Bayesian Dictionary Learning -- A Variational
Bayesian Algorithm}
\end{center}
%\begin{tabular}{p{8cm}}
\vspace{0cm} \noindent
\begin{tabular}{lp{7.7cm}}
\hline 1.& Given the current posterior distributions
$q_{d}(\boldsymbol{D})$, $q_{\alpha}(\boldsymbol{\alpha})$ and
$q_{\gamma}(\gamma)$, update the posterior distribution
$q_{x}(\boldsymbol{X})$ according
to (\ref{x-update}).\\
2.& Given $q_{x}(\boldsymbol{X})$,
$q_{\alpha}(\boldsymbol{\alpha})$, and $q_{\gamma}(\gamma)$,
update
$q_{d}(\boldsymbol{D})$ according to (\ref{D-update}).\\
3.& Given $q_{x}(\boldsymbol{X})$, $q_{d}(\boldsymbol{D})$ and
$q_{\gamma}(\gamma)$, update
$q_{\alpha}(\boldsymbol{\alpha})$ according to (\ref{alpha-update}).\\
4.& Given $q_{x}(\boldsymbol{X})$, $q_{d}(\boldsymbol{D})$ and
$q_{\alpha}(\boldsymbol{\alpha})$, update $q_{\gamma}(\gamma)$
according to
(\ref{gamma-update}).\\
4.& Repeat the above steps until a stopping criterion is reached.\\
\hline
\end{tabular}

\vspace{0.3cm}

\emph{Remarks:} We discuss the choice of the parameter $\beta$
which defines the variance of the dictionary atoms. We might like
to set $\beta$ equal to $1/m$ such that the norm of each atom has
unit variance. Our experiment results, however, suggest that a
very large value of $\beta$, e.g. $10^8$, leads to better
performance. In fact, choosing an infinitely large $\beta$ implies
placing non-informative priors over the atoms
$\{\boldsymbol{d}_n\}$, in which case the update of the dictionary
is simplified as
\begin{align}
\langle\boldsymbol{D}\rangle=\boldsymbol{B}\boldsymbol{A}=\boldsymbol{Y}\langle\boldsymbol{X}^T\rangle
\langle\boldsymbol{X}\boldsymbol{X}^T\rangle^{-1}
\end{align}
This update formula is similar to the formula used for dictionary
update in the MOD method, except with the point estimate
$\boldsymbol{X}$ and $\boldsymbol{X}\boldsymbol{X}^T$ replaced by
the posterior mean $\langle\boldsymbol{X}\rangle$ and
$\langle\boldsymbol{X}\boldsymbol{X}^T\rangle$, respectively.
Nevertheless, unlike the MOD method which alternates between two
separate stages (i.e. dictionary update and sparse coding), for
our algorithm, the dictionary and the signal are refined in an
interweaved and gradual manner, which enables the algorithm to
come to a reasonably nearby point as the optimization progresses,
and helps avoid undesirable local minima. This explains why our
proposed method outperforms the MOD method.

%which requires to compute an $n\times n$ matrix inverse and may
%not be practical for very large number of dictionary columns. To
%address this issue

In the above algorithm, atoms are updated in a parallel way. By
assuming posterior independence among atoms
$\{\boldsymbol{d}_n\}$, our method can also be readily adapted to
update atoms in a sequential manner, i.e. update one atom at a
time while fixing the rest atoms in the dictionary. The mean field
approximation, in this case, can be expressed as
\begin{align}
p(\boldsymbol{\theta}|\boldsymbol{y})\approx &
q(\boldsymbol{x},\boldsymbol{\alpha},\boldsymbol{D},\gamma)\nonumber\\
=& q_x(\boldsymbol{x})q_{\alpha}(\boldsymbol{\alpha})\prod_{n=1}^N
q_{d_n}(\boldsymbol{d}_n)q_{\gamma}(\gamma)
\end{align}
The posterior distribution $q_{d_n}(\boldsymbol{d}_n)$ can then be
computed by maximizing $L(q)$ while keeping other hidden variables
fixed using their most recent distributions, which leads to
\begin{align}
\ln q_{d_n}(\boldsymbol{d}_n)\propto &\langle \ln
p(\boldsymbol{Y},\boldsymbol{X},\{\boldsymbol{d}_k\},\boldsymbol{\alpha},\gamma)\rangle_{q_x(\boldsymbol{X})
\prod_{k\neq n}^N
q_{d_k}(\boldsymbol{d}_k)q_{\alpha}(\boldsymbol{\alpha})q_{\gamma}(\gamma)}
\nonumber\\
\propto &\langle \ln
p(\boldsymbol{Y}|\boldsymbol{X},\{\boldsymbol{d}_k\},\gamma)p(\boldsymbol{d}_n)\rangle_{q_x(\boldsymbol{X})
\prod_{k\neq n}^N q_{d_k}(\boldsymbol{d}_k)q_{\gamma}(\gamma)}
\nonumber\\
\stackrel{(a)}{\propto} &\langle \ln
p(\boldsymbol{Y}^{-n}|\boldsymbol{d}_n,\boldsymbol{x}_{n\cdot},\gamma)p(\boldsymbol{d}_n)\rangle_{q_x(\boldsymbol{X})
\prod_{k\neq n}^N q_{d_k}(\boldsymbol{d}_k)q_{\gamma}(\gamma)}
\nonumber\\
\stackrel{(b)}{\propto} &
\frac{1}{2}\langle\gamma\text{tr}\{(\boldsymbol{Y}^{-n}-\boldsymbol{d}_n\boldsymbol{x}_{n\cdot})
(\boldsymbol{Y}^{-n}-\boldsymbol{d}_n\boldsymbol{x}_{n\cdot})^T\}
\nonumber\\
& +\beta^{-1}\boldsymbol{d}_n^T\boldsymbol{d}_n
\rangle \nonumber\\
=&\frac{1}{2}\left[\boldsymbol{d}_n^T(\langle\gamma\rangle\langle
\boldsymbol{x}_{n\cdot}\boldsymbol{x}_{n\cdot}^T\rangle+\beta^{-1})^{-1}\boldsymbol{d}_n
-2\boldsymbol{d}_n\langle\boldsymbol{Y}^{-n}\rangle\langle
\boldsymbol{x}_{n\cdot}^T\rangle\right] \label{eqn-4}
\end{align}
where in $(a)$, we define
\begin{align}
\boldsymbol{Y}^{-n}\triangleq\boldsymbol{Y}-\boldsymbol{D}^{-n}\boldsymbol{X}
\label{definition-1}
\end{align}
in which $\boldsymbol{D}^{-n}$ is generated by $\boldsymbol{D}$
with the $n$th column of $\boldsymbol{D}$ replaced by a zero
vector, and $\boldsymbol{x}_{n\cdot}$ denotes the $n$th row of
$\boldsymbol{X}$, $(b)$ comes from the fact that
$\boldsymbol{Y}^{-n}-\boldsymbol{d}_n\boldsymbol{x}_{n\cdot}=\boldsymbol{W}$
and thus we have
\begin{align}
p(\boldsymbol{Y}^{-n}|\boldsymbol{d}_n,\boldsymbol{x}_{n\cdot},\gamma)=&
p(\boldsymbol{Y}^{-n}-\boldsymbol{d}_n\boldsymbol{x}_{n\cdot})
\nonumber\\
=&\frac{\gamma^{\frac{ML}{2}}}{2\pi}\exp
\bigg(-\frac{1}{2}\gamma\|\boldsymbol{Y}^{-n}-\boldsymbol{d}_n\boldsymbol{x}_{n\cdot}\|_F^2\bigg)
\label{eqn-5}
\end{align}
From (\ref{eqn-4}), it can be seen that $\boldsymbol{d}_n$ follows
a Gaussian distribution
\begin{align}
q_{d_n}(\boldsymbol{d}_n)=\mathcal{N}(\boldsymbol{d}_n|\boldsymbol{\mu}_n^d,
\boldsymbol{\Sigma}_{n}^d)
\end{align}
with the mean and the covariance matrix given respectively by
\begin{align}
\boldsymbol{\mu}_n^d=&\boldsymbol{\Sigma}_n^d\langle\boldsymbol{Y}^{-n}\rangle\langle
\boldsymbol{x}_{n\cdot}^T\rangle \nonumber\\
\boldsymbol{\Sigma}_{n}^d=& (\langle\gamma\rangle\langle
\boldsymbol{x}_{n\cdot}\boldsymbol{x}_{n\cdot}^T\rangle+\beta^{-1})^{-1}\boldsymbol{I}
\end{align}
where $\langle
\boldsymbol{x}_{n\cdot}\boldsymbol{x}_{n\cdot}^T\rangle$ is the
$n$th diagonal element of
$\langle\boldsymbol{X}\boldsymbol{X}^T\rangle$, and
$\langle\boldsymbol{Y}^{-n}\rangle=\boldsymbol{Y}-\langle\boldsymbol{D}^{-n}\rangle\langle\boldsymbol{X}\rangle$.
Our proposed algorithm therefore can be readily extended to a
columnwise update procedure by replacing the update of
$q_d(\boldsymbol{D})$ with the sequential update of
$q_{d_n}(\boldsymbol{d}_n), \forall n$.

%We only need to alter the way of updating $q_d(\boldsymbol{D})$,
%which is elaborated as follows.

%also commonly used as a means of

\section{Gibbs Sampler} \label{sec:Gibbs}
Gibbs sampling is an effective alternative to the variational
Bayes method for Bayesian inference. In particular, different from
the variational Bayes which provides a locally-optimal, exact
analytical solution to an approximation of the posterior, Monte
Carlo techniques such as Gibbs sampling provide a numerical
approximation to the exact posterior of hidden variables using a
set of samples. It has been observed in a series of experiments
(including our results) that the Gibbs sampler provides better
performance than the variational Bayesian inference.
%denoising.

Let
$\boldsymbol{\theta}\triangleq\{\boldsymbol{X},\boldsymbol{\alpha},
\boldsymbol{D},\gamma\}$ denote all hidden variables in our
hierarchical model. We aim to find the posterior distribution of
$\boldsymbol{\theta}$ given the observed data $\boldsymbol{Y}$
\begin{align}
p(\boldsymbol{\theta}|\boldsymbol{Y})\propto
p(\boldsymbol{Y}|\boldsymbol{D},\boldsymbol{X},\gamma)p(\boldsymbol{D})p(\boldsymbol{X}|\boldsymbol{\alpha})
p(\boldsymbol{\alpha})p(\gamma)
\end{align}
To provide an approximation to the posterior distribution of the
hidden variables, the Gibbs sampler generates an instance from the
distribution of each hidden variable in turn, conditional on the
current values of the other hidden variables. It can be shown
(see, for example, \cite{Gelman13}) that the sequence of samples
constitutes a Markov chain, and the stationary distribution of
that Markov chain is just the sought-after joint distribution.
Specifically, the sequential sampling procedure of the Gibbs
sampler is given as follows.
\begin{itemize}
\item Sampling $\boldsymbol{X}$ according to its conditional
marginal distribution
$p(\boldsymbol{X}|\boldsymbol{Y},\boldsymbol{D}^{(t)},\boldsymbol{\alpha}^{(t)},\gamma^{(t)})$;
\item Sampling $\boldsymbol{D}$ according to its conditional
marginal distribution
$p(\boldsymbol{D}|\boldsymbol{Y},\boldsymbol{X}^{(t+1)},\boldsymbol{\alpha}^{(t)},\gamma^{(t)})$;
\item Sampling $\boldsymbol{\alpha}$ according to its conditional
marginal distribution
$p(\boldsymbol{\alpha}|\boldsymbol{Y},\boldsymbol{D}^{(t+1)},\boldsymbol{X}^{(t+1)},\gamma^{(t)})$;
\item Sampling $\gamma$ according to its conditional
marginal distribution
$p(\gamma|\boldsymbol{Y},\boldsymbol{D}^{(t+1)},\boldsymbol{X}^{(t+1)},\boldsymbol{\alpha}^{(t+1)})$.
\end{itemize}
Note that the above sampling scheme is also referred to as a
blocked Gibbs sampler \cite{Bishop07} because it groups two or
more variables together and samples from their joint distribution
conditioned on all other variables, rather than sampling from each
one individually. Details of this sampling scheme are provided
below. For simplicity, the notation $p(\boldsymbol{z}|-)$ is used
in the following to denote the distribution of variable
$\boldsymbol{z}$ conditioned on all other variables.

%Gibbs sampling is applicable when the joint distribution is not
%known explicitly or is difficult to sample from directly, but the
%conditional distribution of each variable is known and is easy (or
%at least, easier) to sample from.

%Gibbs sampling is a simple and widely applicable Markov chain
%Monte Carlo algorithm.

\textbf{\emph{1). Sampling $\boldsymbol{X}$}}: The samples of
$\boldsymbol{X}$ can be obtained by independently sampling each
column of $\boldsymbol{X}$, i.e. $\boldsymbol{x}_l$. The
conditional marginal distribution of $\boldsymbol{x}_l$ is given
as
\begin{align}
p(\boldsymbol{x}_l|-)&\propto
p(\boldsymbol{Y}|\boldsymbol{X},\boldsymbol{D},\gamma)p(\boldsymbol{x}_l|\boldsymbol{\alpha}_{l})
\nonumber\\
&\propto
p(\boldsymbol{y}_l|\boldsymbol{D},\boldsymbol{x}_l,\gamma)p(\boldsymbol{x}_l|\boldsymbol{\alpha}_{l})
\end{align}
Recalling (\ref{eqn-1}), it can be easily verified that
$p(\boldsymbol{x}_l|-)$ follows a Gaussian distribution
\begin{align}
p(\boldsymbol{x}_l|-)=\mathcal{N}(\boldsymbol{\mu}_{l}^x,\boldsymbol{\Sigma}_{l}^x)
\label{x-gibbs-sampling}
\end{align}
with its mean $\boldsymbol{\mu}^{x}_l$ and covariance matrix
$\boldsymbol{\Sigma}^{x}_l$ given by
\begin{align}
\boldsymbol{\mu}^{x}_l
&=\gamma\boldsymbol{\Sigma}^{x}_l\boldsymbol{D}^T\boldsymbol{y}_l\\
\boldsymbol{\Sigma}^{x}_l&=(\gamma\boldsymbol{D}^T\boldsymbol{D}+\boldsymbol{\Lambda}_l)^{-1}
\end{align}
where
$\boldsymbol{\Lambda}_l\triangleq\text{diag}(\alpha_{1l},\ldots,\alpha_{Nl})$.

\textbf{\emph{2). Sampling $\boldsymbol{D}$}}: There are two
different ways to sample the dictionary: we can sample the whole
set of atoms at once, or sample the atoms in a successive way.
Here, in order to expedite the convergence of the Gibbs sampler,
we sample the atoms of the dictionary in a sequential manner. The
conditional distribution of $\boldsymbol{d}_n$ can be written as
\begin{align}
p(\boldsymbol{d}_n|-)&\propto p(\boldsymbol{d}_n)p(\boldsymbol{Y}|\boldsymbol{D},\boldsymbol{X},\gamma) \nonumber\\
&\propto
p(\boldsymbol{d}_n)p(\boldsymbol{Y}^{-n}|\boldsymbol{d}_n,\boldsymbol{x}_{n\cdot},\gamma)
\end{align}
where $\boldsymbol{Y}^{-n}$ is defined in (\ref{definition-1}).
Recalling (\ref{eqn-5}), we can show that the conditional
distribution of $\boldsymbol{d}_n$ follows a Gaussian distribution
\begin{align}
p(\boldsymbol{d}_n|-)=
\mathcal{N}(\boldsymbol{\mu}_{n}^d,\boldsymbol{\Sigma}_{n}^d)
\label{d-gibbs-sampling}
\end{align}
with its mean and covariance matrix given by
\begin{align}
\boldsymbol{\mu}_{n}^d&=\gamma\boldsymbol{\Sigma}_n^d\boldsymbol{Y}^{-n}\boldsymbol{x}_{n\cdot}^T\\
\boldsymbol{\Sigma}_n^d&=(\gamma\boldsymbol{x}_{n\cdot}
\boldsymbol{x}_{n\cdot}^T+\beta^{-1})^{-1}\boldsymbol{I}
\end{align}

\textbf{\emph{3). Sampling $\boldsymbol{\alpha}$}}: The
log-conditional distribution of $\alpha_{nl}$ can be computed as
\begin{align}
\ln p(\alpha_{nl}|-) &\propto \ln p(\alpha_{nl}; a,b)p(x_{nl}|\alpha_{nl}) \nonumber \\
&\propto \left(a-\frac{1}{2}\right)\ln\alpha_{nl}-\left(b+\frac{
x_{nl}^2}{2}\right)
\end{align}
It is easy to verify that $\alpha_{nl}$ still follows a Gamma
distribution
\begin{align}
p(\alpha_{nl}|-)= \text{Gamma}(\hat{a},\hat{b}_{nl})
\label{alpha-gibbs-sampling}
\end{align}
with the parameters $\hat{a}$ and $\hat{b}_{nl}$ given as
\begin{align}
\hat{a}&=a+\frac{1}{2}\\
\hat{b}_{nl}&=b+\frac{1}{2}x_{nl}^{2}
\end{align}

\textbf{\emph{4). Sampling $\gamma$}}: The log-conditional
distribution of $\gamma$ is given by
\begin{align}
\ln p(\gamma|-) &\propto \ln p(\boldsymbol{Y}|\boldsymbol{D},\boldsymbol{X},\gamma)p(\gamma) \nonumber\\
&\propto \ln\prod_{l=1}^L
p(\boldsymbol{y}_l|\boldsymbol{D},\boldsymbol{x}_l,\gamma)p(\gamma)
\nonumber\\
&=\bigg(\frac{ML}{2}+c-1\bigg)\ln\gamma-
\bigg(\frac{1}{2}\|\boldsymbol{Y}-\boldsymbol{D}\boldsymbol{X}\|_F^2+d\bigg)\gamma
\end{align}
from which we can arrive at
\begin{align}
p(\gamma|-)=
\text{Gamma}(\hat{c},\hat{d})\label{gamma-gibbs-sampling}
\end{align}
where
\begin{align}
\hat{c}&=a+\frac{ML}{2}\\
\hat{d}&=d+\frac{1}{2}\|\boldsymbol{Y}-\boldsymbol{D}\boldsymbol{X}\|_{F}^2
\end{align}

So far we have derived the conditional marginal distributions for
hidden variables
$\{\boldsymbol{D},\boldsymbol{X},\boldsymbol{\alpha},\gamma\}$.
Gibbs sampler successively generates the samples of these
variables according to their conditional distributions. After a
burn-in period, the generated samples can be viewed as samples
drawn from the posterior distribution
$p(\boldsymbol{X},\boldsymbol{D},
\boldsymbol{\alpha},\gamma|\boldsymbol{Y})$. With those samples,
the dictionary can be estimated by averaging the last few samples
of the Gibbs sampler. For clarity, we now summarize the Gibbs
sampling algorithm as follows.

\begin{center}
\textbf{Sparse Bayesian Dictionary Learning -- A Gibbs Sampling
Algorithm}
\end{center}
%\begin{tabular}{p{8cm}}
\vspace{0cm} \noindent
\begin{tabular}{lp{7.7cm}}
\hline 1.& Given the current samples $\boldsymbol{D}^{(t)}$,
$\boldsymbol{\alpha}^{(t)}$ and $\gamma^{(t)}$.
Generate a sample $\boldsymbol{X}^{(t+1)}$ according to (\ref{x-gibbs-sampling}).\\
2.& Given the current samples $\boldsymbol{X}^{(t+1)}$,
$\boldsymbol{\alpha}^{(t)}$ and $\gamma^{(t)}$.
Generate a sample $\boldsymbol{D}^{(t+1)}$ according to (\ref{d-gibbs-sampling}).\\
3.& Given the current samples $\boldsymbol{D}^{(t+1)}$,
$\boldsymbol{X}^{(t+1)}$ and $\gamma^{(t)}$.
Generate a sample $\boldsymbol{\alpha}^{(t+1)}$ according to (\ref{alpha-gibbs-sampling}).\\
4.& Given the current samples $\boldsymbol{D}^{(t+1)}$,
$\boldsymbol{X}^{(t+1)}$ and $\boldsymbol{\alpha}^{(t+1)}$.
Generate a sample $\gamma^{(t+1)}$ according to (\ref{gamma-gibbs-sampling}).\\
5.& Repeat the above steps and collect the samples after a burn-in
period. \\
\hline
\end{tabular}

\section{Simulation Results} \label{sec:simulation}
We now carry out experiments to illustrate the performance of our
proposed sparse Bayesian dictionary learning (SBDL) methods, which
are respectively referred to as SBDL-VB and SBDL-Gibbs. Throughout
our experiments, the parameters for our proposed method are set
equal to $a=0.5$, $b=10^{-6}$, $c=0.5$, and $d=10^{-6}$. The
parameter $\beta$ is set to $\beta=10^{8}$ for the SBDL-VB and
$\beta=1$ for SBDL-Gibbs. Note that the SBDL-Gibbs is insensitive
to the choice $\beta$ and here we simply choose $\beta=1$. We
compare our proposed methods with other existing state-of-the-art
dictionary learning methods, namely, the K-SVD algorithm
\cite{AharonElad06}, the atom parallel-updating (APrU-DL) method
\cite{SadeghiZadeh14}, and the Bata-Bernoulli process factor
analysis (BPFA) method \cite{ZhouChen12}. Both the synthetic data
and real data are used to test the performance of respective
algorithms.

%In our simulations, $L$ is set to be $1000$ and $2000$,
%respectively.

\subsection{Synthetic Data}
We generate a dictionary $\boldsymbol{D}$ of size $20\times50$,
with each entry independently drawn from a normal distribution.
Columns of $\boldsymbol{D}$ are then normalized to unit norm. The
training signals $\{\boldsymbol{y}_l\}_{l=1}^L$ are produced based
on $\boldsymbol{D}$, where each signal $\boldsymbol{y}_l$ is a
linear combination of $K_l$ randomly selected atoms and the
weighting coefficients are i.i.d. normal random variables. Two
different cases are considered. First, all training samples are
generated with the same number of atoms, i.e. $K_l=K,\forall l$,
and $K$ is assumed exactly known to the K-SVD method. The other
case is that $K_l$ varies from $3$ to $6$ for different $l$
according to a uniform distribution. In this case, the K-SVD
assumes that the sparsity level equals to $6$ during the sparse
coding stage. The observation noise is assumed multivariate
Gaussian with zero mean and covariance matrix
$\sigma^2\boldsymbol{I}$. Note that the APrU-DL (with FISTA)
method requires to set two regularization parameters $\lambda$ and
$\lambda_s$ to control the tradeoff between the sparsity and the
data fitting error. The selection of these two parameters is
always a tricky issue and an inappropriate choice may lead to
considerable performance degradation. To show this, we use the
following two different choices:
$\{\lambda,\lambda_s\}=\{0.2,0.15\}$ and
$\{\lambda,\lambda_s\}=\{0.4,0.4\}$, in which the former set of
values are carefully selected to achieve the best performance, and
the latter set of values slightly deviate from the former set of
values. We use APrU-DL-F to denote the APrU-DL method which uses
the former choice, and APrU-DL-L to denote the APrU-DL method
which uses the latter one. For SBDL-Gibbs, the number of
iterations is set to $300$ and the estimate of the dictionary is
simply chosen to be the last sample of the Gibbs sampler. For a
fair comparison, the competing algorithms including K-SVD,
APrU-DL, and BPFA are executed sufficient numbers of iterations to
achieve their best performance.

The recovery success rate is used to evaluate the dictionary
learning performance. The success rate is computed as the ratio of
the number of successfully recovered atoms to the total number of
atoms. An atom is considered successfully recovered if the
distance between the original atom and the estimated atom is
smaller than 0.01, where the distance is defined as
\begin{align}
1-\frac{|\boldsymbol{d}_i^T\hat{\boldsymbol{d}}_i|}{\|\boldsymbol{d}_i\|\|\hat{\boldsymbol{d}}_i\|}
\end{align}
where $\hat{\boldsymbol{d}_i}$ denotes the estimated atom. Table
\ref{table1} shows the average recovery success rates of
respective algorithms, where we set $L=1000$ and $L=2000$,
respectively, and the signal-to-noise ratio (SNR) varies from 10
to 100dB. Results are averaged over 50 independent trials. From
Table \ref{table1}, we can see that:
\begin{itemize}
\item The proposed SBDL-Gibbs method achieves the highest recovery success rates in most
cases. The proposed SBDL-VB method, although not as well as the
SBDL-Gibbs, still provides quite decent performance and presents a
clear performance advantage over the K-SVD and APrU-DL methods
when the number of training signals is limited, e.g. $L=1000$. In
particular, both the SBDL-Gibbs and the SBDL-VB outperform the
BPFA method by a big margin, although all these three methods were
developed in a Bayesian framework.
\item In the low SNR regime, e.g.
$\text{SNR}=10\text{dB}$, the K-SVD method suffers from a
significant performance loss when there is a discrepancy between
the presumed sparsity level and the groundtruth (see the case
where $K_l$ varies but the presumed sparsity level is fixed to 6).
\item The APrU-DL method is sensitive to the choice of the regularization parameters.
It provides superior recovery performance when the regularization
parameters are properly selected. Nevertheless, as we can see from
Table \ref{table1}, the APrU-DL method incurs a considerable
performance degradation when the parameters deviate from their
optimal choice, and there is no general guideline suggesting how
to choose appropriate values for these regularization parameters.
\end{itemize}

%performs similarly to the K-SVD and APrU-DL methods when $L=2000$,
%while

\begin{table}
\caption{Recovery Success Rates}
\begin{tabular}{|c|c|c|c|c|c|c|}
    \hline
    $L$ & SNR & Algorithm & $K$ = 3 & $K$ = 4 & $K$ = 5 & Var. $K$\\
    \hline
    \hline
    \multirow{24}{*}{1000}
    &\multirow{6}{*}{10}
    & K-SVD & 80.52 & 36.36 & 2.52 & 0.80 \\
    \cline{3-7}
    && BPFA & 64.48 & 38.00 & 11.60 & 26.56\\
    \cline{3-7}
    && APrU-DL-F & 85.64 & \textbf{64.40} & \textbf{33.44} & \textbf{53.68}\\
    \cline{3-7}
    && APrU-DL-L & 48.20 & 17.48 & 4.68 & 12.52\\
    \cline{3-7}
    && SBDL-VB & 86.00 & 63.84 & 16.28 & 47.48\\
    \cline{3-7}
    && SBDL-Gibbs & \textbf{91.52} & 62.48 & 6.32 & 41.80\\
%    \hline
%    \hline
    \cline{2-7}
    &\multirow{6}{*}{20}
    & K-SVD & 93.20 & 93.44 & 92.08 & 84.68 \\
    \cline{3-7}
    && BPFA & 83.20 & 85.88 & 85.00 & 85.08\\
    \cline{3-7}
    && APrU-DL-F & 94.04 & 93.32 & 87.76 & 93.48\\
    \cline{3-7}
    && APrU-DL-L & 72.48 & 40.32 & 14.15 & 33.04\\
    \cline{3-7}
    && SBDL-VB & 97.28 & 95.96 & 92.32 & 94.48\\
    \cline{3-7}
    && SBDL-Gibbs & \textbf{99.64} & \textbf{99.16} & \textbf{97.52} & \textbf{99.12}\\
%    \hline
%    \hline
    \cline{2-7}
    &\multirow{6}{*}{30}
    & K-SVD & 94.24 & 94.32 & 93.92 & 86.64 \\
    \cline{3-7}
    && BPFA & 75.72 & 80.68 & 82.96 & 81.56\\
    \cline{3-7}
    && APrU-DL-F & 94.24 & 94.92 & 88.16 & 93.96\\
    \cline{3-7}
    && APrU-DL-L & 73.40 & 43.16 & 17.16 & 34.36\\
    \cline{3-7}
    && SBDL-VB & 96.60 & 96.16 & 92.32 & 95.48\\
    \cline{3-7}
    && SBDL-Gibbs & \textbf{99.60} & \textbf{99.16} & \textbf{98.64} & \textbf{99.00}\\
%    \hline
%    \hline
    \cline{2-7}
    &\multirow{6}{*}{100}
    & K-SVD & 94.24 & 94.32 & 93.92 & 85.44 \\
    \cline{3-7}
    && BPFA & 75.88 & 78.96 & 82.16 & 78.24\\
    \cline{3-7}
    && APrU-DL-F & 94.36 & 93.84 & 88.68 & 93.64\\
    \cline{3-7}
    && APrU-DL-L & 74.88 & 44.72 & 18.12 & 36.08\\
    \cline{3-7}
    && SBDL-VB & 97.20 & 97.52 & 92.24 & 94.56\\
    \cline{3-7}
    && SBDL-Gibbs & \textbf{99.32} & \textbf{99.24} & \textbf{98.24} & \textbf{98.96}\\
    \hline
%\end{tabular}
%\label{tab:1}
%\end{table}
%
%
%\begin{table}
%\caption{Success Rate of Recovery ($L$ = 2000)}
%\begin{tabular}{|c|c|c|c|c|c|c|}
%    \hline
%    L & SNR (dB) & Algorithm & $k$ = 3 & $k$ = 4 & $k$ = 5 & unfixed $k$ \\
%    \hline
    \hline
    \multirow{24}{*}{2000}
    & \multirow{6}{*}{10}
    & K-SVD & 91.00 & 88.88 & 50.56 & 25.32 \\
    \cline{3-7}
    && BPFA & 85.44 & 82.84 & 67.92 & 81.24\\
    \cline{3-7}
    && APrU-DL-F & 97.00 & 94.88 & \textbf{86.24} & \textbf{95.44}\\
    \cline{3-7}
    && APrU-DL-L & 84.84 & 68.36 & 42.28 & 64.04\\
    \cline{3-7}
    && SBDL-VB & 92.92 & 81.80 & 55.68 & 77.16\\
    \cline{3-7}
    && SBDL-Gibbs & \textbf{98.56} & \textbf{95.72} & 80.20 & 93.88\\
%    \hline
%    \hline
    \cline{2-7}
    & \multirow{6}{*}{20}
    & K-SVD & 95.64 & 96.68 & 95.16 & 94.00 \\
    \cline{3-7}
    && BPFA & 84.44 & 87.16 & 88.48 & 86.68\\
    \cline{3-7}
    && APrU-DL-F & 95.40 & 96.48 & 95.80 & 96.56\\
    \cline{3-7}
    && APrU-DL-L & 85.32 & 82.44 & 64.48 & 79.84\\
    \cline{3-7}
    && SBDL-VB & 97.64 & 96.56 & 92.12 & 95.04\\
    \cline{3-7}
    && SBDL-Gibbs & \textbf{99.48} & \textbf{99.56} & \textbf{98.92} & \textbf{99.16}\\
%    \hline
%    \hline
    \cline{2-7}
    & \multirow{6}{*}{30}
    & K-SVD & 95.88 & 96.92 & 96.96 & 93.36 \\
    \cline{3-7}
    && BPFA & 76.84 & 81.00 & 83.60 & 81.64\\
    \cline{3-7}
    && APrU-DL-F & 94.28 & 95.00 & 96.80 & 95.64\\
    \cline{3-7}
    && APrU-DL-L & 86.32 & 82.40 & 66.08 & 80.52\\
    \cline{3-7}
    && SBDL-VB & 96.88 & 96.96 & 92.96 & 94.96\\
    \cline{3-7}
    && SBDL-Gibbs & \textbf{99.40} & \textbf{99.16} & \textbf{99.52} & \textbf{99.32}\\
%    \hline
%    \hline
    \cline{2-7}
    & \multirow{6}{*}{100}
    & K-SVD & 96.04 & 97.88 & 96.88 & 92.20 \\
    \cline{3-7}
    && BPFA & 75.24 & 80.12 & 83.00 & 81.36\\
    \cline{3-7}
    && APrU-DL-F & 95.56 & 95.48 & 96.08 & 96.00\\
    \cline{3-7}
    && APrU-DL-L & 86.36 & 82.92 & 65.76 & 79.56\\
    \cline{3-7}
    && SBDL-VB & 96.84 & 96.56 & 94.32 & 96.36\\
    \cline{3-7}
    && SBDL-Gibbs & \textbf{99.40} & \textbf{99.44} & \textbf{99.24} & \textbf{99.40}\\
    \hline
\end{tabular}
\label{table1}
\end{table}

%carry out experiments on natural images

\subsection{Application To Image Denoising}
We now demonstrate the results by applying the above methods to
image denoising. Suppose images are corrupted by white Gaussian
noise with zero mean and variance $\sigma^2$. We partition a
noise-corrupted image into a number of overlapping patches of size
$8\times 8$ pixels. Note that in our simulations, not all patches
are selected for training, but only those patches whose top-left
pixels are located at $[r\times i,r\times j]$ for any
$i,j=0,\ldots,\lfloor(Q-8)/r\rfloor$ are selected, where $Q$
denotes the dimension of the $Q\times Q$ image, and $r$ is chosen
to be $r=\{2,4\}$, respectively. The selected patches are then
vectorized to generate the training signal $\{\boldsymbol{y}_l\}$.
Also, in our experiments, we assume that the noise variance is
perfectly known \emph{a priori} by the K-SVD method. For the
APrU-DL method, the regularization parameters $\lambda$ and
$\lambda_s$ are carefully chosen to be $\lambda=25$ and
$\lambda_s=30$. After the training by respective algorithms, the
trained dictionary is then used for denoising. The denoising
process involves a sparse coding of all patches (including those
used for training and those not) of size $8\times 8$ pixels from
the noisy image. Due to its simplicity and fast execution, the
orthogonal matching pursuit (OMP) method is employed to perform
the sparse coding of all patches. The final estimate of each pixel
is obtained by averaging the associated pixel from each of the
denoised overlapping patches in which this pixel is included.

\begin{figure*}[htb]
    \centering
    \includegraphics [width=100pt]{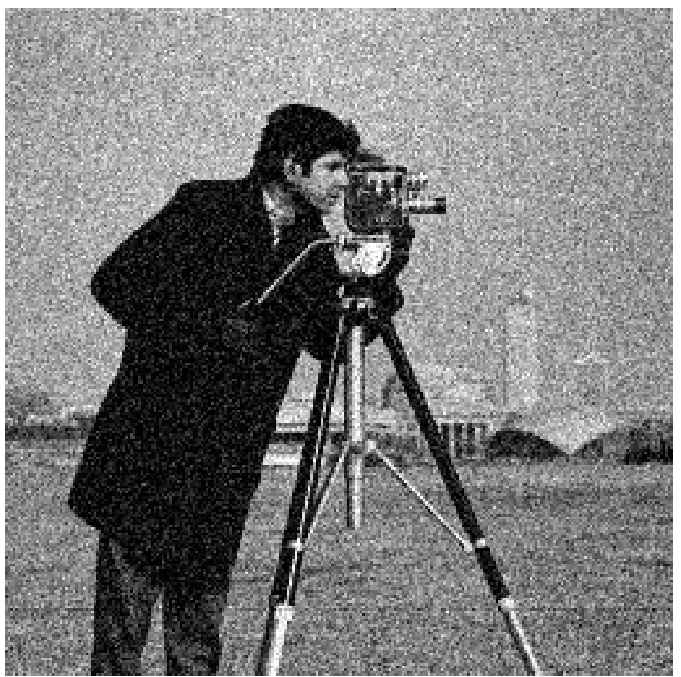}
    \includegraphics [width=100pt]{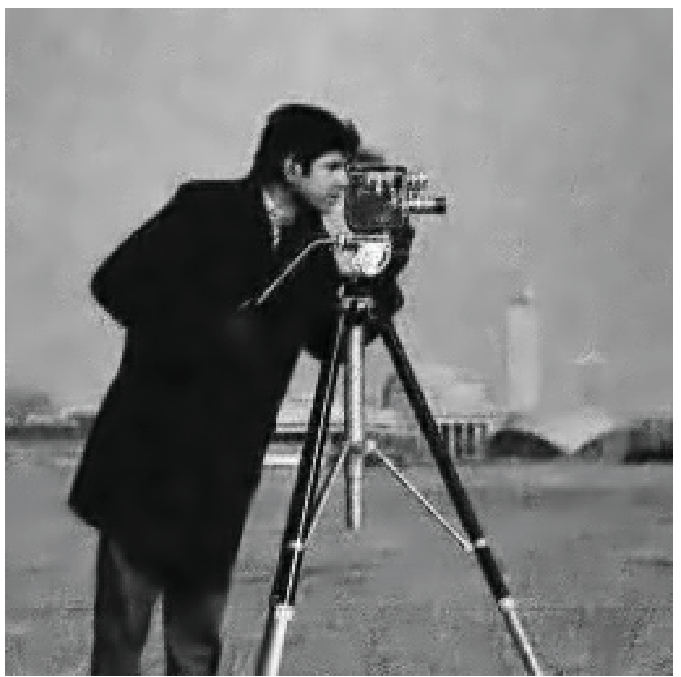}
    \includegraphics [width=100pt]{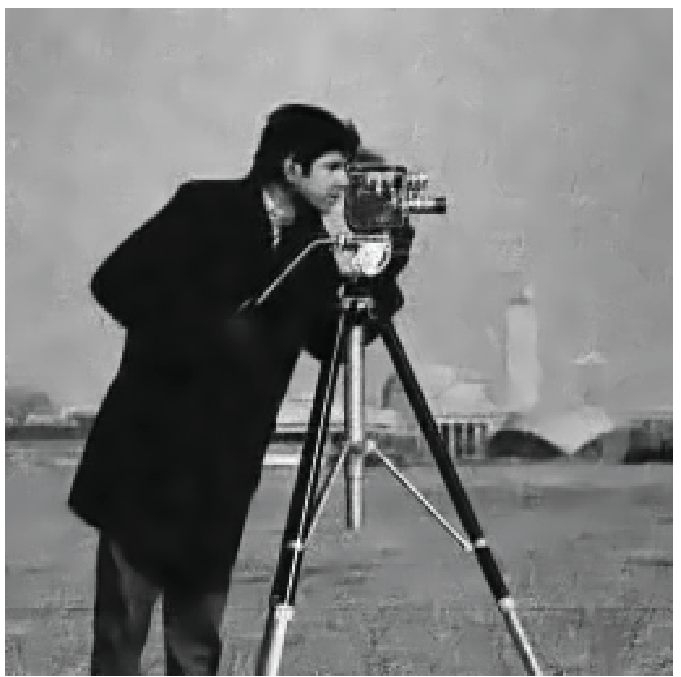}
    \includegraphics [width=100pt]{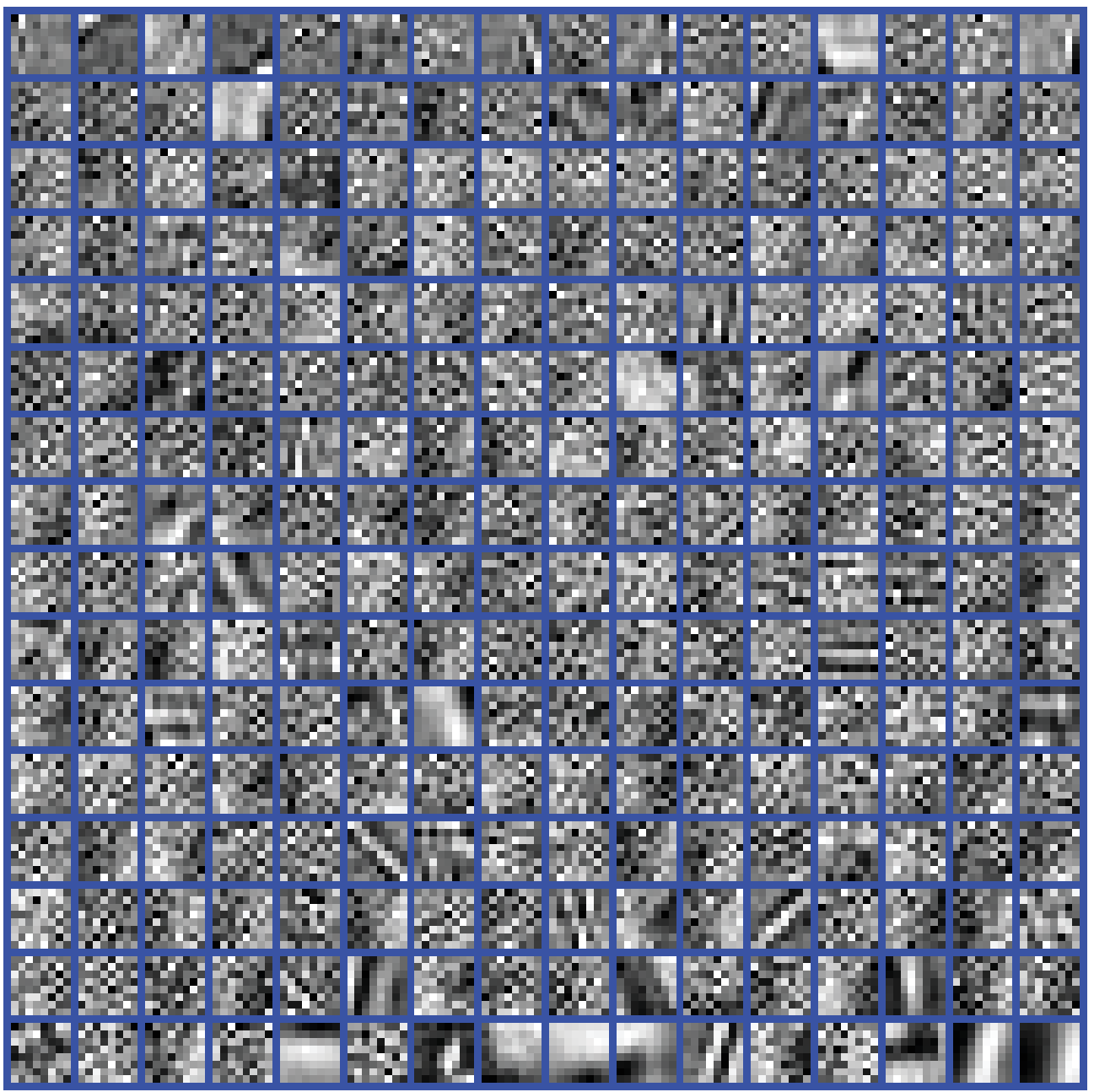}
    \includegraphics [width=100pt]{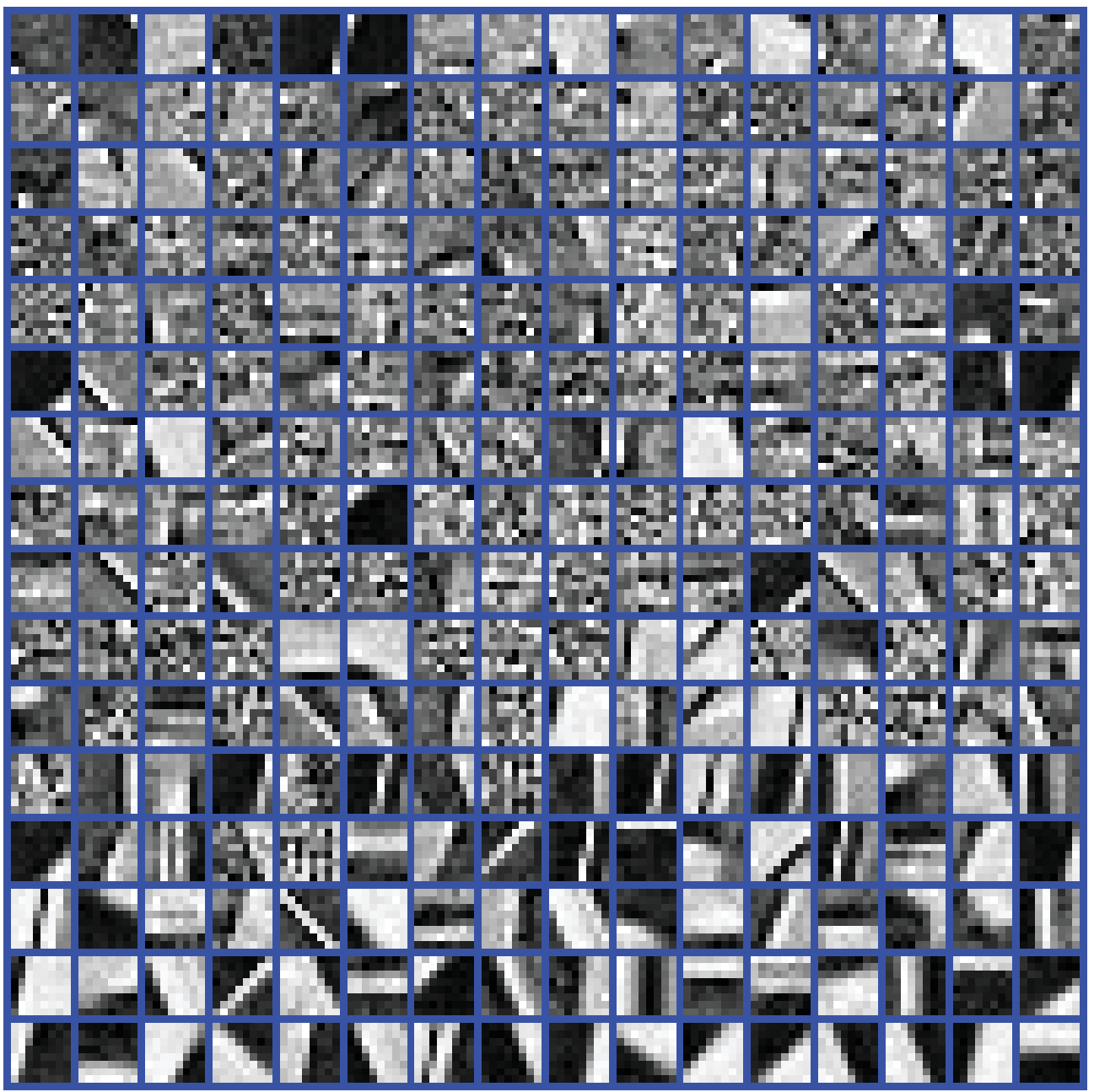}
    \caption{Example of the denoising results for the image ``Cameraman'' ($\sigma=25$ and $r$=2). (From left to right)
    The corrupted image, the denoised image by SBDL-VB (28.1587dB), the denoised image
by SBDL-Gibbs (28.8380dB), the dictionary trained by SBDL-VB, the
dictionary trained by SBDL-Gibbs.}
    \label{fig:camera}
\end{figure*}

\begin{figure*}[htb]
    \centering
    \includegraphics [width=100pt]{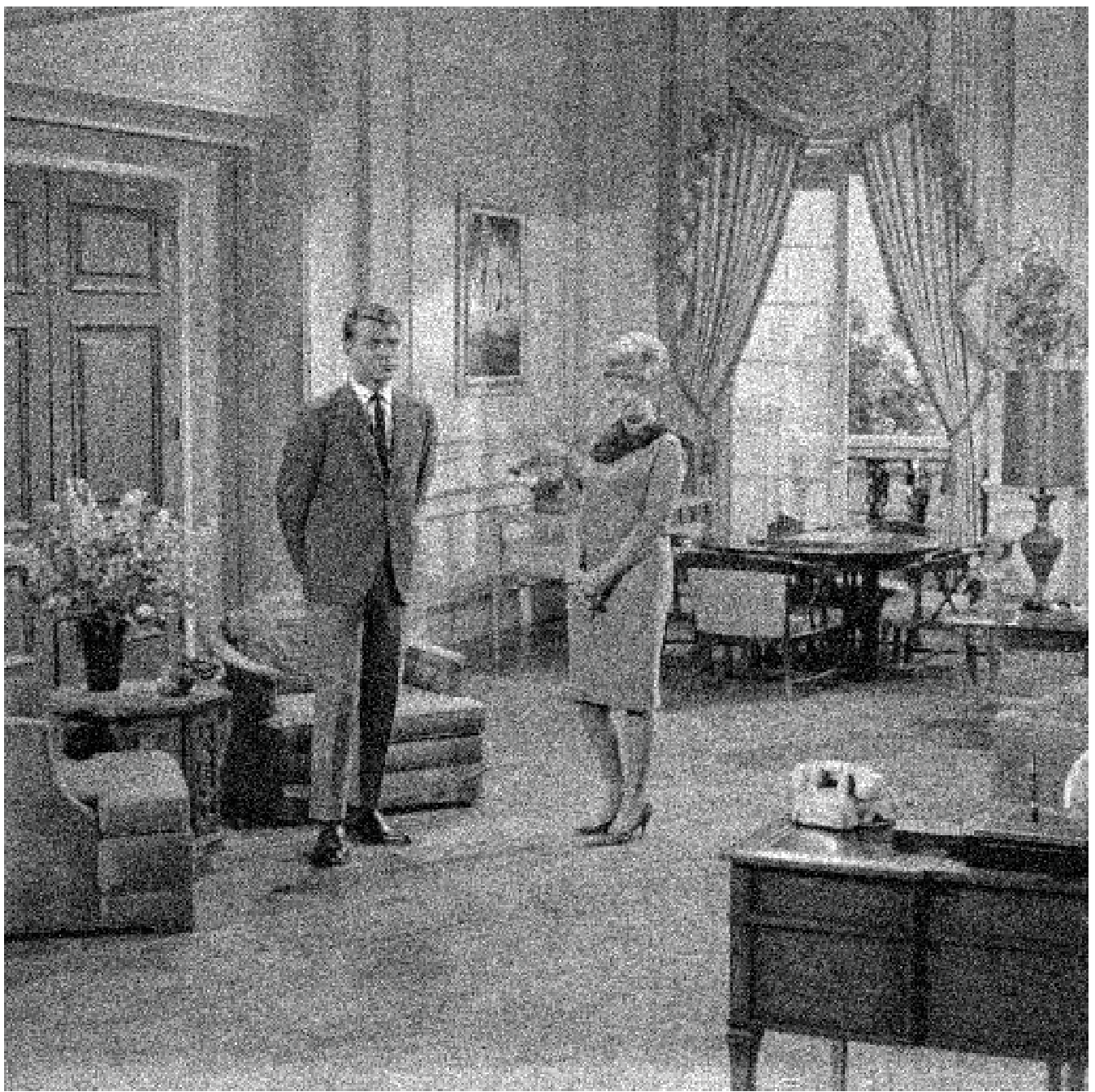}
    \includegraphics [width=100pt]{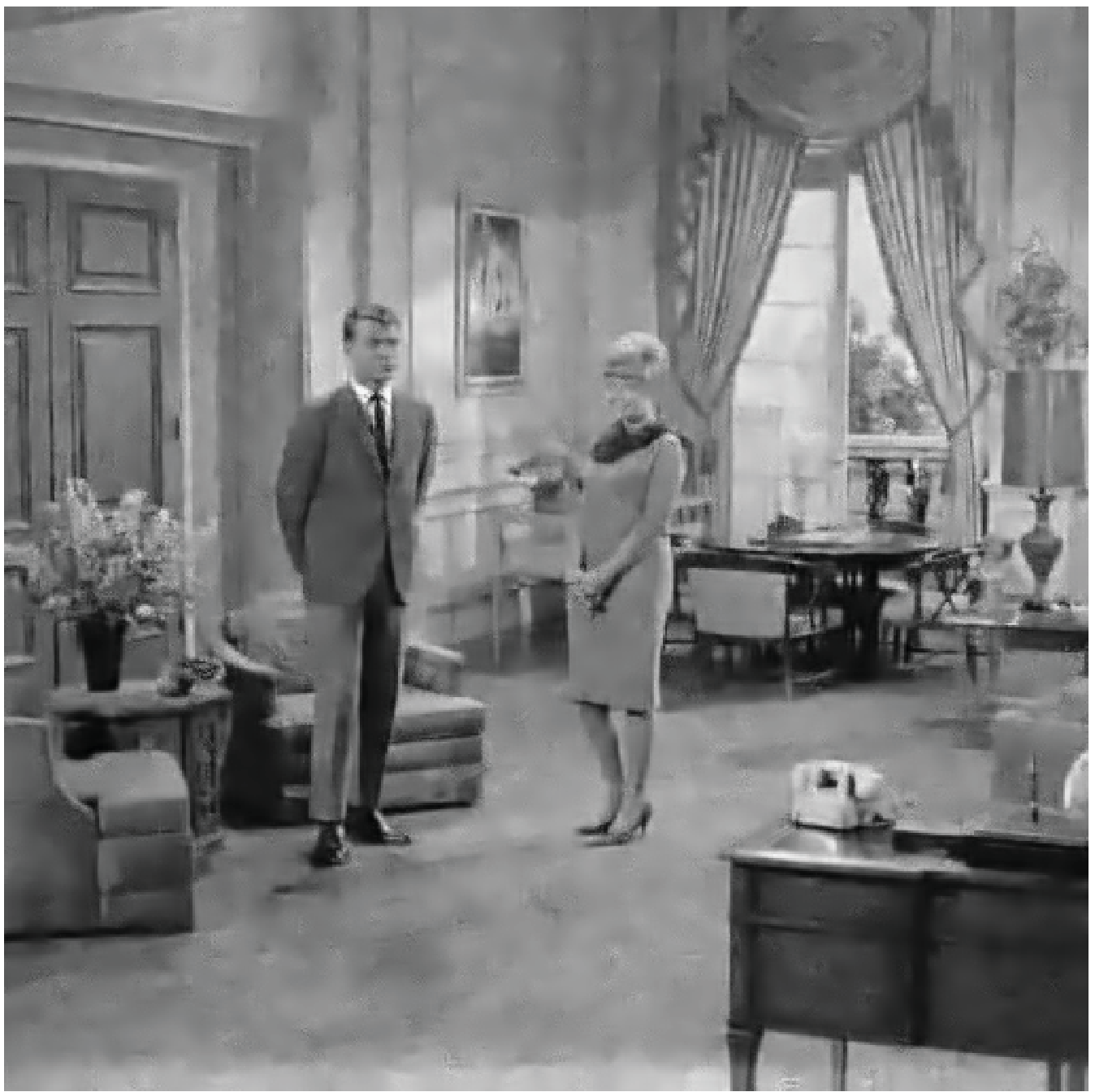}
    \includegraphics [width=100pt]{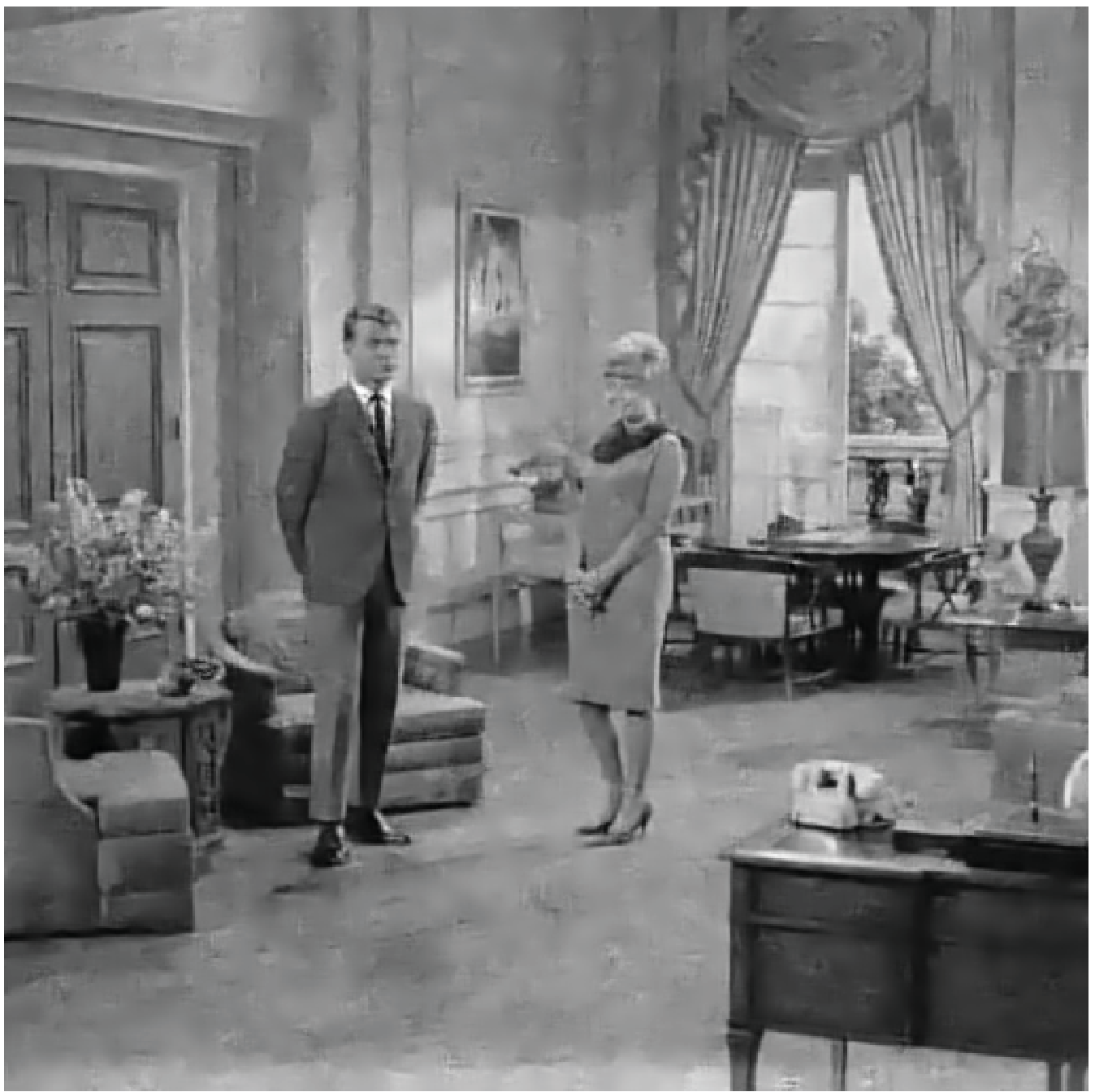}
    \includegraphics [width=100pt]{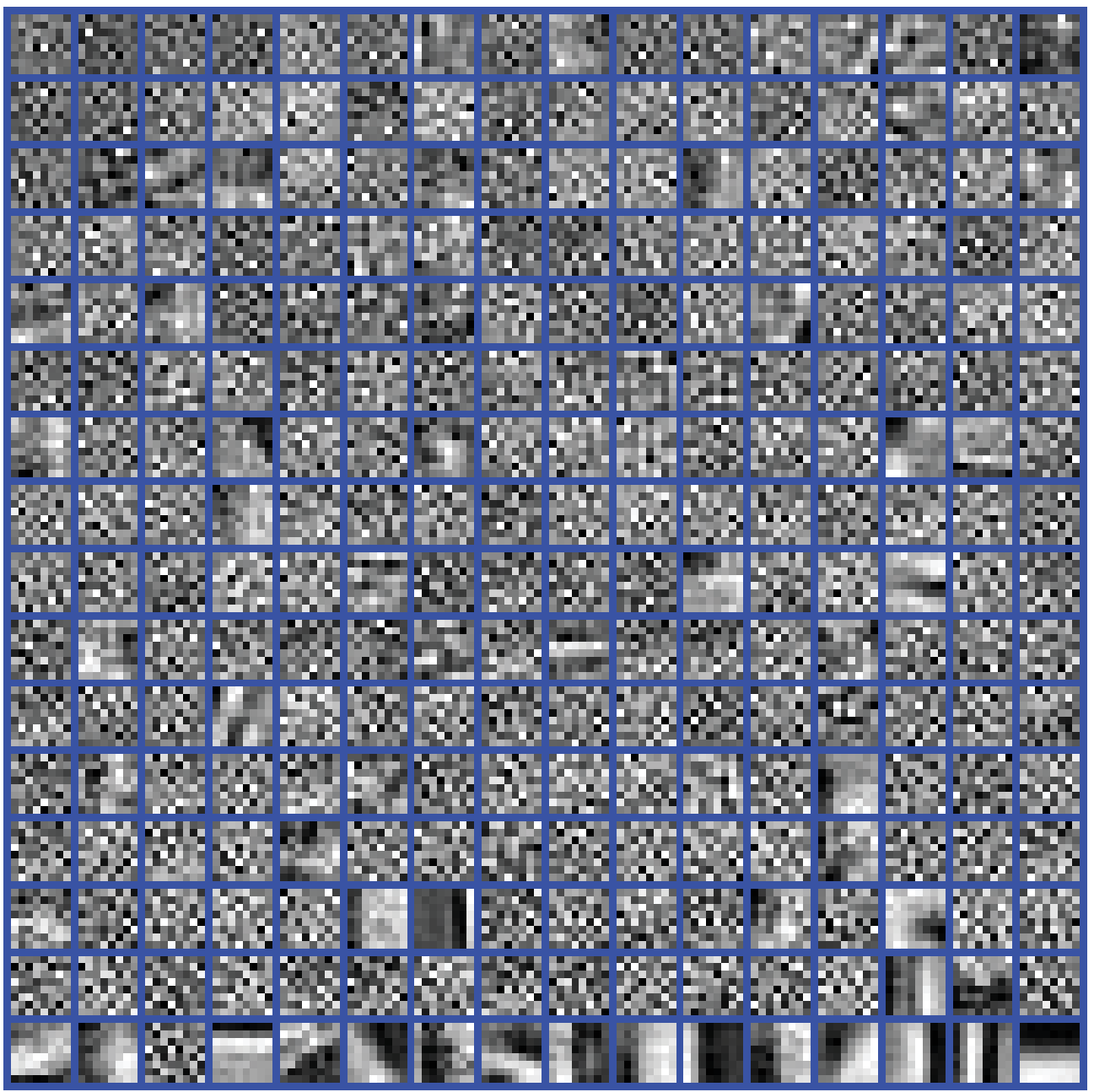}
    \includegraphics [width=100pt]{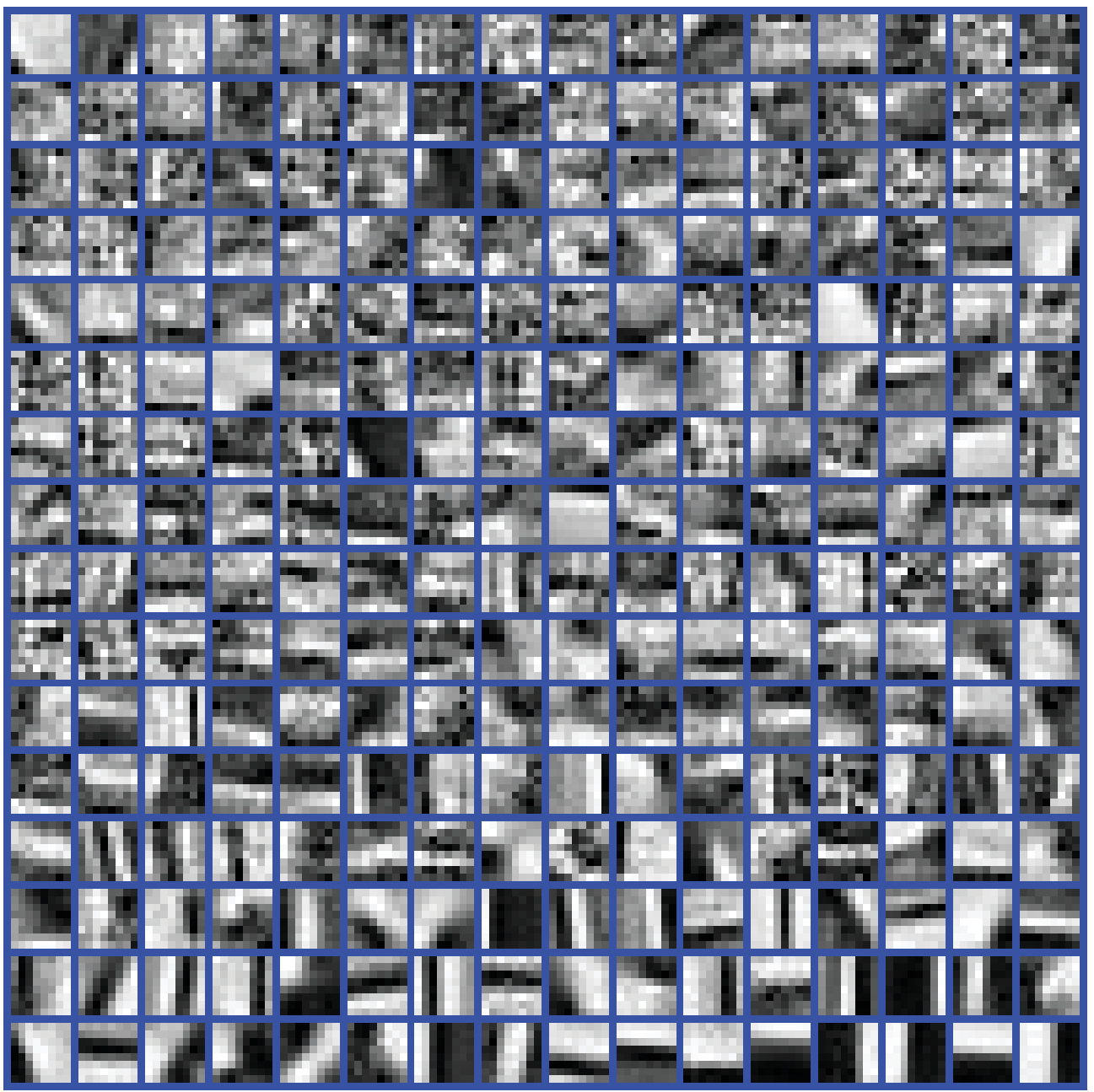}
    \caption{Example of the denoising results for the image ``Couple'' ($\sigma=25$ and $r$=2). (From left to right)
    The corrupted image, the denoised image by SBDL-VB (28.4240dB), the denoised image
    by SBDL-Gibbs (28.8431dB), the dictionary trained by SBDL-VB, the dictionary trained by SBDL-Gibbs.}
    \label{fig:couple}
\end{figure*}

Table \ref{table2} shows the peak signal to noise ratio (PSNR)
results obtained for different nature images by respective
algorithms, where the noise standard deviation is set to
$\sigma=\{15,25,50\}$, respectively, and the dictionary to be
inferred is assumed of size $64\times 256$. The PSNR is defined as
\[
    \text{PSNR} = 20\log_{10}\bigg(\frac{255\times Q^2}{\|\boldsymbol{\hat{U}}-\boldsymbol{U}\|_F}\bigg)
\]
where $\boldsymbol{\hat{U}}$ and $\boldsymbol{U}$ denote the
denoised image and the original image, respectively. From Table
\ref{table2}, we see that the results of all methods are very
close to each other in general. The proposed SBDL-Gibbs achieves a
slightly higher PSNR than other methods in most cases,
particularly when less number of signals is used for training.
This result again demonstrates the superiority of the proposed
method. In Fig. \ref{fig:camera} and \ref{fig:couple}, we present
the noise-corrupted images ``cameraman'' and ``couple'', and the
denoised images using dictionaries trained by our proposed
algorithms. The trained dictionaries are also shown on the right
sides of Fig. \ref{fig:camera} and \ref{fig:couple}.

\begin{table}
\caption{PSNR Results}
\begin{center}
\begin{tabular}{|c|c|c|c|c|c|}
    \hline
    r & $\sigma$ & Algorithm & boat & cameraman & couple \\
    \hline
    \hline
     \multirow{15}{*}{2}
    & \multirow{5}{*}{15}
    & K-SVD & 29.2802 & 31.4638 & 31.4068 \\
    \cline{3-6}
    && BPFA & 29.2988 & 30.8684 & 31.0950 \\
    \cline{3-6}
    && APrU-DL & 29.5718 & \textbf{31.7662} & \textbf{31.5304} \\
    \cline{3-6}
    && SBDL-VB & 29.3557 & 31.1741 & 31.0691 \\
    \cline{3-6}
    && SBDL-Gibbs & \textbf{29.5881} & 31.6978 & 31.4473 \\
    \cline{2-6}
    & \multirow{5}{*}{25}
    & K-SVD & 26.9308 & 28.6211 & 28.6949 \\
    \cline{3-6}
    && BPFA & 26.9576 & 28.1639 & 28.5825 \\
    \cline{3-6}
    && APrU-DL & 26.8998 & 28.7069 & 28.5378 \\
    \cline{3-6}
    && SBDL-VB & 26.6959 & 28.1587 & 28.4240 \\
    \cline{3-6}
    && SBDL-Gibbs & \textbf{27.1570} & \textbf{28.8380} & \textbf{28.8431} \\
    \cline{2-6}
    & \multirow{5}{*}{50}
    & K-SVD & 22.9499 & 23.9898 & 24.3532 \\
    \cline{3-6}
    && BPFA & \textbf{23.5059} & 22.8861 & 24.3181 \\
    \cline{3-6}
    && APrU-DL & 22.7274 & 23.5888 & 24.1901 \\
    \cline{3-6}
    && SBDL-VB & 23.0861 & 23.3194 & 24.3299 \\
    \cline{3-6}
    && SBDL-Gibbs & 23.4651 & \textbf{24.1899} & \textbf{24.7870} \\
    \hline
    \hline
     \multirow{15}{*}{4}
    & \multirow{5}{*}{15}
    & K-SVD & 29.2585 & 31.3553 & 31.3513 \\
    \cline{3-6}
    && BPFA & 28.8131 & 29.7561 & 30.3464 \\
    \cline{3-6}
    && APrU-DL & 29.4554 & \textbf{31.5541} & 31.4276 \\
    \cline{3-6}
    && SBDL-VB & 29.3217 & 31.0739 & 31.1359 \\
    \cline{3-6}
    && SBDL-Gibbs & \textbf{29.5376} & 31.4931 & \textbf{31.5443} \\
    \cline{2-6}
    & \multirow{5}{*}{25}
    & K-SVD & 26.6756 & 28.4350 & 28.5580 \\
    \cline{3-6}
    && BPFA & 26.3727 & 26.8885 & 27.5139 \\
    \cline{3-6}
    && APrU-DL & 26.7240 & 28.4447 & 28.4097 \\
    \cline{3-6}
    && SBDL-VB & 26.5977 & 28.0960 & 28.3715 \\
    \cline{3-6}
    && SBDL-Gibbs & \textbf{27.0077} & \textbf{28.5539} & \textbf{28.7889} \\
    \cline{2-6}
    & \multirow{5}{*}{50}
    & K-SVD & 22.7708 & 23.2908 & 24.2388 \\
    \cline{3-6}
    && BPFA & 23.0100 & 22.0422 & 23.4463 \\
    \cline{3-6}
    && APrU-DL & 22.6036 & 23.3086 & 24.1107 \\
    \cline{3-6}
    && SBDL-VB & 23.0404 & 23.3315 & 24.4163 \\
    \cline{3-6}
    && SBDL-Gibbs & \textbf{23.2525} & \textbf{23.8610} & \textbf{24.6326} \\
    \hline
\end{tabular}
\end{center}
\label{table2}
\end{table}


%Benefited from the carefully chosen parameters, the APrU-DL method
%also demonstrates superior denoising results.

%such that the best denoising performance is achieved.

%Different choices of $r=\{2,4\}$ and $\sigma=\{15,25,50\}$ are
%considered in our simulations. The dictionary to be inferred is
%assumed of size $64\times 256$.

\section{Conclusions} \label{sec:conclusion}
We developed a new Bayesian hierarchical model for learning the
overcomplete dictionaries based on a set of training data. This
new framework can be considered as an adaptation of the
conventional sparse Beysian learning framework to deal with the
dictionary learning problem. Specifically, a Gaussian-inverse
Gamma hierarchical prior is used to promote the sparsity of the
representation. Suitable priors are also placed on the dictionary
and the noise variance such that they can be reasonably inferred
from the data. We developed a variational Bayesian method and a
Gibbs sampler for Bayesian inference. Unlike some of previous
methods, the proposed methods do not need to assume knowledge of
the noise variance \emph{a priori}, and can infer the noise
variance automatically from the data. The performance of the
proposed methods is evaluated using synthetic data. Numerical
results show that the proposed methods are able to learn the
dictionary with an accuracy considerably better than existing
methods, particularly for the case where there is a limited number
of training signals. The proposed methods are also applied to
image denoising, where superior denoising results are achieved
even compared to other state-of-the-art algorithms. Our proposed
hierarchical model is also flexible to incorporate additional
prior information to enhance the dictionary learning performance.
%This will a topic of our future study.

%\bibliography{newbib}
%\bibliographystyle{IEEEtran}

\end{document}